\documentclass[10pt,twocolumn,letterpaper]{article}

\usepackage{iccv}
\usepackage{times}
\usepackage{epsfig}
\usepackage{graphicx}
\usepackage{amsmath}
\usepackage{amssymb}
\usepackage{booktabs}  
\usepackage{mathtools}
\usepackage[table]{xcolor}

\makeatletter
\@namedef{ver@everyshi.sty}{}
\makeatother
\usepackage{nicematrix}

\usepackage[pagebackref=true,breaklinks=true,letterpaper=true,colorlinks,bookmarks=false]{hyperref}

\iccvfinalcopy 

\def\iccvPaperID{****} 

\ificcvfinal\fi

\usepackage{times,epsfig,graphicx}
\usepackage{algorithm,algorithmic}
\usepackage{verbatim}
\usepackage{tabularx}
\usepackage{xspace}
\usepackage{bbm}
\usepackage{bm}

\usepackage{amssymb,amsmath}
\newlength{\smallimage}
        \definecolor{rel}{rgb}{.1,.6,.2}
        \definecolor{nrl}{rgb}{1,1,1}
        \definecolor{qim}{rgb}{1,1,1}

\makeatletter

\def\eg{\emph{e.g.\,}}

\def\ie{\emph{i.e.\,}}

\def\etc{\emph{etc.\,}}
\def\wrt{w.r.t.\,}

\def\iid{i.i.d.\,}
\def\etal{\emph{et al.\,}}

\newcommand{\fig}[1]{Figure~\ref{fig:#1}}
\newcommand{\tabl}[1]{Table~\ref{tab:#1}}

\def\be{\begin{equation}}
\def\ee{\end{equation}}
\def\bea{\begin{eqnarray}}
\def\eea{\end{eqnarray}}
\def\ben{\begin{eqnarray*}}
\def\een{\end{eqnarray*}}

\def\bi{\begin{itemize}}
\def\ei{\end{itemize}}

\newcommand{\btab}[1]{\begin{tabular}{#1}}
\newcommand{\etab}{\end{tabular}}
\newcommand{\ba}[1]{\begin{array}{#1}}
\newcommand{\ea}{\end{array}}

\def\Loss{\mathcal L}

\DeclareMathOperator*{\argmax}{\mathrm{argmax}}

\def\<{\langle}
\def\>{\rangle}

\newcommand{\bfx}{{\bf x}}
\newcommand{\bfy}{{\bf y}}

\newcommand{\calD}{{\mathcal D}}

\newcommand{\calS}{{\mathcal S}}
\newcommand{\calT}{{\mathcal T}}

\newcommand{\calX}{{\mathcal X}}
\newcommand{\calY}{{\mathcal Y}}

\newcommand{\Exp}{\mathbb{E}}

\newcommand{\myparagraph}[1]{\vspace{0.1cm}\noindent\textbf{#1.}}



\definecolor{DarkCoral}{rgb}{0.8, 0.36, 0.27}

\begin{document}

\title{Unsupervised Domain Adaptation for Semantic Image Segmentation:\\ a Comprehensive Survey}

\author{Gabriela Csurka,  Riccardo Volpi and Boris Chidlovskii\\\
Naver Labs Europe,  France \\
\url{https://europe.naverlabs.com}\\
{\tt\small firstname.lastname@naverlabs.com}
}

\maketitle
\ificcvfinal\thispagestyle{empty}\fi

\begin{abstract}
Semantic segmentation plays a fundamental role in a broad variety of computer vision applications, providing key information for the global understanding of an image. Yet, the state-of-the-art models rely on large amount of annotated samples, which are more expensive to obtain than in 
tasks such as image classification. Since unlabelled data is instead significantly cheaper to obtain, it is not surprising that Unsupervised Domain Adaptation reached a broad success within the semantic segmentation community.

This survey is an effort to summarize five years of this incredibly rapidly growing field, which embraces the importance of semantic segmentation itself and a critical need of adapting 
segmentation models to new environments. We present the most important semantic segmentation methods; we provide a comprehensive survey on domain adaptation techniques for semantic segmentation; we unveil newer trends such as multi-domain learning, domain generalization, test-time adaptation or source-free domain adaptation; we 
conclude this survey by describing datasets and benchmarks most widely used in semantic segmentation research. 
We hope that this survey will provide researchers across academia and industry with a comprehensive reference guide and will help them in fostering new research directions in the field.
\end{abstract}

\section{Introduction}\label{sec:intro}

Semantic image segmentation (SiS) plays a fundamental role in a general understanding of the image content and context. It aims to label image pixels with the corresponding semantic classes and to provide boundaries of the class objects, 
easing the understanding of
object appearances and the spatial relationships between them. Therefore, it represents an important task 
towards the design of artificial intelligent systems that will fuel applications in fields of autonomous driving~\cite{TeichmannIVS18MultiNetRealTimeJointSemReasoningAD,HofmarcherBC19VisualSceneUnderstandingAutonomousDriving}, 
medical image analysis~\cite{RonnebergerMICCAI15UNetSegmentation}, 
augmented reality~\cite{DeChicchisTR20SemanticUnderstandingAR,TurkmenTR19SceneUnderstandingThroughSemSegmAR}, 
satellite imaging~\cite{MaJPRS19DeepLearningRemoteSensingApplications}, etc. 

Recent advances in deep learning allowed a significant performance boost in many computer vision tasks, including semantic segmentation. However, the success of deep learning methods typically depends on the availability of large amounts of annotated training data. Manual annotation of images with pixel-wise semantic labels is an extremely tedious and time consuming process. Initially taking an hour or more per image~\cite{CordtsCVPR16CityscapesDataset}, recent semi-automatic tools\footnote{See as example the \url{https://github.com/errolPereira/Semi-Automatic-Image-Annotation-Tool}} manage to reduce the annotation time for common urban classes (people, road surface or vehicles) by relying, \eg, on pre-trained models for object detection, however they still require manual verification and validation\footnote{We refer to the dedicated page \url{https://github.com/heartexlabs/awesome-data-labeling} for an up-to-date collection of annotation tools.}.

Progress in computer graphics and modern high-level generic graphics platforms, such as game engines, enabled the generation of photo-realistic virtual worlds with diverse, realistic, and physically plausible events and actions. The computer vision and machine learning communities realized that such tools can be used to generate datasets for training deep learning models~\cite{RichterECCV16PlayingForData}. Indeed, such synthetic rendering pipelines can produce a virtually unlimited amount of labeled data --- leading to good performance when deploying models on real data, due to constantly increasing photorealism of the rendered datasets. Furthermore, it becomes easy to diversify data generation; for example, when generating scenes that simulate driving conditions, one can simulate seasonal, weather, daylight or architectural style changes, making such data generation pipeline suitable to support the design and training of computer vision models for diverse tasks, such as SiS.

While some SiS models trained on simulated images can already perform relatively well on real images, their performance can be further improved by domain adaptation (DA) --- and in particular {\it unsupervised domain adaptation} (UDA) --- that can bridge the gap caused by the domain shift between the synthetic and real images. For the aforementioned reasons, sim-to-real adaptation represents one of the leading benchmarks to assess the effectiveness of domain adaptation for semantic image segmentation (DASiS).

In contrast to most DA surveys~\cite{CsurkaX20DeepVisualDomainAdaptation,WangNC18DeepVisualDASurvey,wilson20surveyUDADeep}, which address generic approaches by mainly covering image classification and only briefly focusing on few adaptation methods for semantic segmentation,
our survey is centered on the DASiS literature.  
Some SiS adaptation problems are also covered in recent surveys on domain generalization~\cite{WangX20GeneralizingToUnseenDomainsSurveyDG,ZhouX20DomainGeneralizationSurvey}, online learning~\cite{HoiX18OnlineLearningComprehensiveSurvey}
and robot perception~\cite{GargFTR20SemanticsForRoboticMappingPerceptionIInteractionSurvey}.

The survey most similar to ours is by Toldo~\etal~\cite{ToldoX20UnsupervisedDASSReview},
sharing the same goal of reviewing the recent advances of unsupervised DASiS. Nevertheless, we believe that our survey extends and enriches~\cite{ToldoX20UnsupervisedDASSReview} in multiple ways; 
in particular we provide the reader with four main contributions, 
which can be summarized as follows:
\begin{itemize}
    \item We review the {\it semantic image segmentation} (SiS) task itself, and cover the existing approaches proposed in the literature;
    \item We organize DASiS methods according to the most {\it important characteristics}, such as the backbone segmentation network, the type and levels of alignment, network sharing, and whether they rely on complementary techniques such as self-training, etc. This is summarized in~\tabl{semsegm_methods}, which represents one of our core contributions.
    \item We go beyond the pure DASiS task and propose a detailed categorization of some {\it related tasks} --- such as domain generalization and source-free adaptation --- and survey the methods addressing them.
    \item We present an extensive comparison of the {\it existing datasets} and discuss how one could use them to test new DASiS methods. We also discuss different evaluation measures and evaluation protocols allowing to fairly compare the different methods proposed in the literature. 
\end{itemize}

The rest of this survey is organized as follows. In Section~\ref{sec:semsegm}, we present the most widely used SiS models, paying  special attention to deep neural networks, including recent visual transformers.
In Section~\ref{sec:dass}, we present and categorize
a large number of approaches devised to  tackle the DASiS task, then we survey complementary techniques that can help boosting adaptation performance; this section
represents the core of our survey. 
In Section~\ref{sec:beyond_dass} we survey the case of 
tasks related to DA --- \eg, multi-source/multi-target domain adaptation, domain generalization, source-free domain adaptation   --- and the methods that address them, still focusing on SiS. 

In Section~\ref{sec:benchmarks} we detail the datasets and benchmarks that practitioners typically use to evaluate adaptation approaches in semantic segmentation; we also cover the main metrics that practitioners rely on to evaluate 
SiS models and discuss  DASiS
evaluation protocols. We draw our final remarks in Section~\ref{sec:conclusions}.

\section{Semantic image segmentation (SiS)}
\label{sec:semsegm}

Semantic image segmentation (SiS) is a computer vision problem 
where the task is determining
to which class each pixel of an image belongs to. This problem is often approached in a supervised learning fashion, by relying on a dataset of images annotated at pixel level, and training a machine learning model to perform this task. The term {\em ``semantic image segmentation''} reflects the goal of determining the nature, \ie semantics, of different parts of an image. A related but different problem is that of low-level image segmentation (not addressed in this survey), which consists in an unsupervised partitioning an image into coherent regions according to some low-level cues, such as color, texture or depth. 

Deep learning-based SiS methods form multiple groups, often around their main principles~~\cite{MinaeeX20ImageSegmUsingDeepLearningSurvey}. Below, we briefly recall the most popular approaches; as the SiS literature is extremely rich, we refer the reader to the above mentioned survey. 

\begin{figure}[ttt]
\begin{center}
\includegraphics[width=0.4\textwidth]{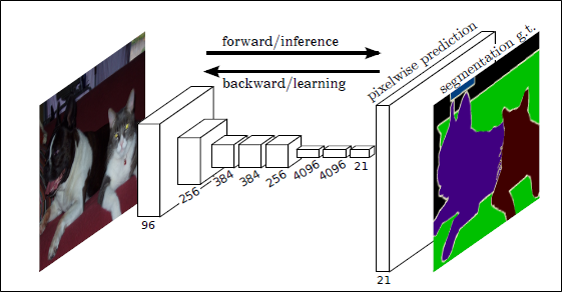}
\caption{The Fully Convolutional Network\cite{LongCVPR15FullyConvolutionalNetworksSegmentation}
that takes an arbitrary size image and produces the same size output, 
is suitable for spatially dense prediction tasks including SiS.}
\label{fig:FCN}
\end{center}
\end{figure}

\myparagraph{Fully convolutional networks (FCNs)} 
The model introduced by Long~\etal~\cite{LongCVPR15FullyConvolutionalNetworksSegmentation} is the first deep network for SiS; it uses fully convolutional layers to process images of arbitrary size
and to produce a segmentation map of the same size (see Figure~\ref{fig:FCN}). 
ParseNet~\cite{LiuX15ParseNetLookingWiderToSeeBetter} 
overcomes some of 
the limitations of FCNs,
by complementing 
these models
with a global context.

\myparagraph{Graphical Models} 
Integrating a larger context in SiS is inspired by the shallow image segmentation methods, and several methods proposed to complement convolutional networks with Conditional Random Fields~\cite{ChandraECCV16FastExactMultiScaleInferenceSemSegmDeepGaussianCRFs,ChenICLR15SemanticImgSegmFullyConnectedCRFs,SchwingX15FullyConnectedDeepStructuredNetworks,ZhengICCV15ConditionalRandomFieldsAsRNN}, 
jointly training the 
CNN parameters
and the fully connected CRF. Teichmann~\etal~\cite{TeichmannBMVC19ConvolutionalCRFsSemSegm} reformulate
CRF inference in terms of convolutions; this allows to improve the efficiency of CRFs, which are known for being hard to optimize and slow at inference time.


\myparagraph{Recurrent Neural Networks}
Another group of methods consider using Recurrent Neural Network (RNN) instead of CNNs; it was shown that modeling the long distance dependencies among pixels is beneficial to improve the segmentation quality. In ReNet, proposed by Visin~\etal~\cite{VisinCVPRWS16ReSegRNNSemSegm}, each 
layer is composed of four RNNs. In a similar way,
Byeon~\etal~\cite{ByeonCVPR15SceneLabelingLSTM} introduce two-dimensional
LSTM networks, which take into account complex spatial dependencies of
labels. Liang~\etal~\cite{LiangECCV16SemanticObjectParsingGraphLSTM} propose
the Graph LSTM model, which 
considers
an arbitrary-shaped superpixel as a semantically
consistent node of the graph and spatial relations between the superpixels as
its edges.

\myparagraph{Pyramid Network Based Models}
Another group of methods generalizes the idea of the Feature Pyramid Network~\cite{LinCVPR17FeaturePyramidNetworksObjDet} proposed for object detection, for semantic segmentation. 
Ghias~\etal~\cite{GhiasiECCV16LaplacianPyramidReconstructionRefinementSemSegm} develop a multi-resolution reconstruction architecture based on a Laplacian pyramid that uses skip connections from higher resolution feature maps and multiplicative gating to successively refine segment boundaries reconstructed from lower-resolution maps. The pyramid scene parsing network (PSPNet)~\cite{ZhaoCVPR17PyramidSceneParsingNetwork} extends the pixel-level features to the specially designed global pyramid pooling one, where the local and global clues are used together to make the final prediction more reliable. He~\etal~\cite{HeCVPR19AdaptivePyramidContextNetworkSemSegm} introduce multi-scale contextual representations with multiple adaptive context modules, where each of such modules uses a global representation to guide the local affinity coefficients estimation for each sub-region.

\myparagraph{Attention-Based Models}
Chen~\etal~\cite{ChenCVPR16AttentionToScaleScalAwarSemSegm} propose an attention mechanism that weigh multi-scale features at each pixel location; furthermore, Li~\etal~\cite{LiBMVC18PyramidAttentionNetworkSemSegm} use Feature Pyramid Attention modules to embed different scale context features in an FCN based pixel prediction framework. Fu~\etal~\cite{FuCVPR19DualAttentionNetworkSceneSegm} introduce Dual Attention Networks to adaptively integrate local features with their global dependencies.

\myparagraph{Encoder-decoder based models} 
These  architectures are composed by an encoder part where the input image is compressed into a latent-space representation that captures the underlying semantic information and a decoder that generates  a predicted output from this latent representation. The model is trained  by minimizing a reconstruction loss  between the GT and the predicted segmentation map~
\cite{BadrinarayananPAMI17SegnetDeepConvEncoderDecoder,NohICCV15LearningDeconvolutionNNSegmentation,RonnebergerMICCAI15UNetSegmentation}
(see,~\eg Figure~\ref{fig:FCN}). To maintain high-resolution representations through the encoding process, several papers consider using HRNet~\cite{SunCVPR19DeepHighResolutionReprHumanPoseEstimation} as backbone instead of ResNet or VGG, since it enables connecting the high-to-low resolution convolution streams in parallel~\cite{SunX19HighResolutionReprLabelingPixelsRegions,YuanECCV20ObjectContextualReprSemSegm}.

\begin{figure}[ttt]
\begin{center}
\includegraphics[width=0.47\textwidth]{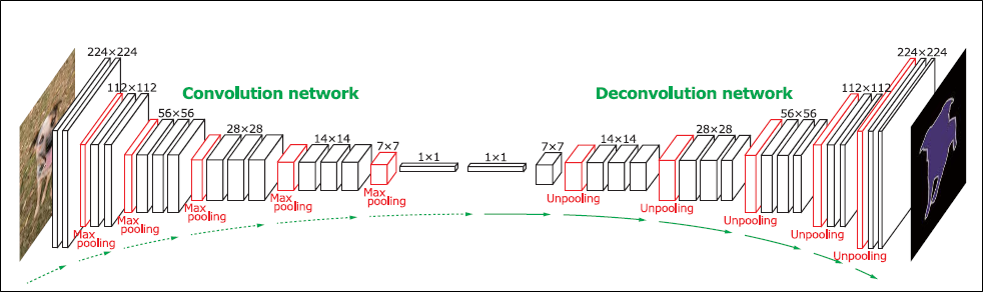}
\caption{The Deconvolutional Network for SiS~\cite{NohICCV15LearningDeconvolutionNNSegmentation} is composed by a multi-layer convolutional network as encoder and a multi-layer deconvolution network as decoder. The latter is build on top of the output of 
the convolutional network, where a series of unpooling, deconvolution and rectification operations is applied to yield a dense pixel-wise class prediction map.}
\label{fig:DeconvNet}
\end{center}
\end{figure}

\myparagraph{Dilated convolutional models} 
Dilated convolutions have been introduced to  improve
the segmentation by multi-scale processing and many recent methods rely on them such as context aggregation~\cite{YuX15MultiScaleContextAggregationDilatedConvolutions}, dense upsampling convolution and hybrid dilated convolution (DUC-HDC)~\cite{WangWACV18UnderstandingConvolutionSemSegm}, densely connected Atrous Spatial Pyramid Pooling (DenseASPP)~\cite{YangCVPR18DenseASPPSemSegmStreetScenes}, and the efficient neural network (ENet)~\cite{PaszkeX16ENetDeepNNForRealTimeSemSegm}.
The most popular is the DeepLab family~\cite{ChenPAMI17DeeplabSemanticImgSegmentationDeepFullyConnectedCRF,ChenX17RethinkingAtrousConvolutionSemSegm,ChenICLR15SemanticImgSegmFullyConnectedCRFs}, which combines several ingredients such as dilated convolution to address the decreasing resolution, Atrous Spatial Pyramid Pooling (ASPP), to capture objects as well as image context at multiple scales, and CRFs to improve the segmentation boundaries. 
Chen~\etal~\cite{ChenECCV18EncoderDecoderAtrousSeparableConvSemSegm} use the DeepLabv3 framework as encoder in an encoder-decoder architecture, termed ``DeepLabv3+''. We note that DeepLabv2~\cite{ChenICLR15SemanticImgSegmFullyConnectedCRFs} remains the most popular network in the DASiS literature.

\myparagraph{Transformer models\footnote{At the moment of submitting this survey, we are unaware of any DASiS solution that relies on transformers; we nevertheless opted to discussing them here, 
given the rapid progress of this sub-field.}} 
This is the most recent effort to capture global image context and to address the segmentation ambiguity at the level of image patches. 
Strudel~\etal~\cite{StrudelICCV21SegmenterTransformerForSemSegm} extend the recent Vision Transformer (ViT) models~\cite{Dosovitskiy2021ImageisWorth16x16Words} to handle semantic segmentation problems. In contrast to convolution-based approaches, it allows to model global context at the first layer and throughout the network. It relies on the output embeddings corresponding to image patches and obtains class labels from these embeddings with a point-wise linear decoder or a mask transformer decoder.
Xie~\etal~\cite{XieX21SegFormerSemSegmTransformers} combine hierarchical transformer-based encoders to extract coarse and fine features with lightweight multilayer perceptron (MLP) decoders that aggregate information from different layers, thus combining both local and global attentions to render a more powerful representation.
Most recently, Guo~\etal~\cite{GuoICCV21SOTRSegmentingObjectsWithTransformers} also follow the hierarchical approach: they use Feature Pyramid Network (FPN) to generate multi-scale feature maps which are then fed into a transformer, to acquire global dependencies and to predict per-instance category, and into a multi-level upsampling module to dynamically generate segmentation masks guided by the transformer output.
Ranftl~\etal~\cite{RanftlICCV21VisionTransformers4DensePrediction} 
introduce a transformer for dense prediction, including both depth estimation and semantic segmentation: the model takes region-wise output of convolutional networks augmented with positional embedding, assembles tokens from various stages of the vision transformer into image-like representations at various resolutions and progressively combines them into full-resolution predictions using a convolutional decoder.

\begin{figure}[ttt]
\begin{center}
\includegraphics[width=0.48\textwidth]{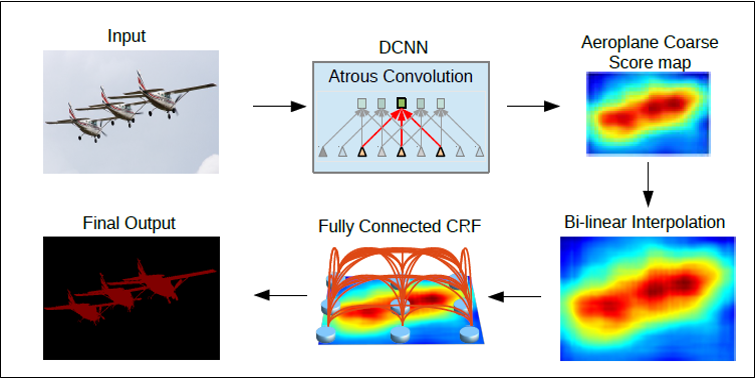}
\caption{The DeepLab model~\cite{ChenICLR15SemanticImgSegmFullyConnectedCRFs} relies on a deep CNN with atrous convolutions to reduce the degree of signal downsampling and a bilinear interpolation stage that enlarges the feature maps to the original image resolution. A fully connected CRF is finally applied to refine the segmentation result and to better capture the object boundaries.}
\label{fig:DeepLab}
\end{center}
\end{figure}


\section{Domain adaptation for semantic image segmentation (DASiS)}
\label{sec:dass}

As we mentioned in Section~\ref{sec:intro}, 
the UDA formulation is particularly intriguing for 
SiS, where data annotation is particularly expensive --- while deep learning based solutions generally rely on large amounts of annotated data. Following the success of UDA techniques 
on deep networks 
for image classification~\cite{GaninJMLR16DomainAdversarialNN}, many deep DA methods have been proposed for SiS as well. 

As major bottleneck in SiS is the high cost of manual annotation, many methods rely on graphics platforms and game engines to generate synthetic data and use them as source domain. These simulation platforms allow to promptly label images at pixel level, and also to simulate various environments, weather or seasonal conditions.

The main DASiS goal is ensuring that SiS models perform well on real target data, by leveraging annotated, synthetic and non-annotated real data.
A classical DASiS framework relies on either SYNTHIA~\cite{RosCVPR16SYNTHIADataset} or GTA~\cite{RichterECCV16PlayingForData} dataset as a source, and the real-world dataset Cityscapes~\cite{CordtsCVPR16CityscapesDataset} as a target. Some known exceptions include domain adaptation between medical images~\cite{BermudezChaconISBI18ADomainAdaptiveUNetElectronMicroscopy,PeroneNI19UDAMedicalImSegmSelfEnsembling}, aerial images~\cite{LeeICCV21SelfMutatingNetworkDASSAerialImages}, weather and seasonal conditions changes of outdoor real images~\cite{WulfmeierIROS17AddressingAppearanceChangeAdvDA}, and different Field of View (FoV) images~\cite{GuICCV21PITPositionInvariantTransformForCrossFoVDA}.

Early DASiS methods were directly inspired by adaptation methods invented for image classification~\cite{CsurkaX20DeepVisualDomainAdaptation,WangNC18DeepVisualDASurvey}.
However SiS is a more complex task, as labeling is done on the pixel level. Methods for image classification commonly embed entire images in some latent space and then align source and target data distributions. Directly applying such a strategy to SiS models is sub-optimal, due to the higher dimensionality and complexity of the output space. 
To address this complexity, most DASiS methods take into account the spatial structure and the local image context, act at multiple levels of the segmentation pipeline and often combine multiple techniques.

Therefore, we step away from grouping the DA methods into big clearly distinguishable families as it is done in recent surveys on image classification~\cite{CsurkaX20DeepVisualDomainAdaptation,WangNC18DeepVisualDASurvey}. 
We instead identify a number of critical characteristics of existing DASiS pipelines and categorize the most prominent methods according to them.
From this point of view,~\tabl{semsegm_methods} is one of our major contributions. It is detailed in Section~\ref{sec:dass_siamese}, where we describe the different domain alignment techniques that are applied at input image, feature and output prediction levels. In Section~\ref{sec:dass_improvements}, we describe complementary machine learning strategies that can empower domain alignment and improve the performance of a segmentation model on target images. Before presenting these different methods, in the next section we first formalize the UDA problem and list the most popular approaches addressing it.

\subsection{UDA and DASiS: notations and background} 
\label{sec:UDA}

Let $\calD_\calS = \calX_\calS \times \calY_\calS$ be a set of paired samples and annotations ($\calX_\calS=\{\bfx_i\}_{i=1}^M$ and $\calY_\calS=\{\bfy_i\}_{i=1}^M$, respectively), drawn from a source distribution $P_\calS(\calX,\calY)$. In the SiS context, $\bfx$ and $\bfy$ represent images and their pixel-wise annotations, respectively, $\bfx \in \mathcal{R}^{H \times W \times 3}$ and $\bfy \in \mathcal{R}^{H \times W \times C}$, where $(H, W)$ is the image size and $C$ is the number of categories. Let $\calD_\calT = \calX_\calT = \{\bfx_i\}_{i=1}^N$ be a set of unlabelled samples drawn from a target distribution $P_\calT$, such that $P_\calS \neq P_\calT$ due to the domain shift. In the UDA setup, both sets are available at training time ($\calD = \calD_\calS \cup \calD_\calT$) and its goal is to learn a model performing well on samples from the target distribution.

Furthermore,  often the segmentation network used in DASiS has an encoder(E)-decoder(D) structure (see \fig{AdaptLevels}), the
domain alignment can happen at different levels of the segmentation network, including the output of the  encoder, at various level of the decoder, or even considering the label predictions as features (as discussed later in this section. Also
features can be build at image level, region level or pixel level. 
Therefore, in which follows when we use the notation of $F_S$ and $F_T$) it refers to any of the above source  respectively target feature generator. Note that it is frequent that the feature encoders $F_S$ and $F_T$ share their parameters $\theta_{F_S}$ and $\theta_{F_T}$ --- in this case, we simply refer to them  as $F$ and $\theta_{F}$.

In the following, we will cover some basic components and commonly used losses that constitute the foundation of most UDA and DASiS approaches. 

\myparagraph{Distribution Discrepancy Minimization}
In image classification, one popular approach to address UDA is to minimize the distribution discrepancy between source and target domains in some latent feature space --- that is, the space of the learned visual representation. This space often corresponds to some activation layers; this includes the last layer the classifier is trained on, but other layers can be considered as well.
One popular measure is the empirical Maximum Mean Discrepancy (MMD)~\cite{BorgwardtBI06IntegratingKernelMaximumMeanDiscrepancy}, that is written as 
\begin{align*}
\Loss_{mmd} = 
\left\| \frac{1}{M}  \sum_{ \bfx_s\in \calX_\calS} \phi(F_S(\bfx_s)) - \frac{1}{N}  \sum_{ \bfx_t\in \calX_\calT} \phi(F_T(\bfx_t)) \right\|\;,
\end{align*}
where $\phi$ is the mapping function corresponding to a RKHS kernel 
defined as a mixture of Gaussian kernels.


\myparagraph{Adversarial Training}
An alternative to minimizing the distribution discrepancy between source and target domains is given by adversarial training~\cite{GoodfellowNIPS14GenerativeAdversarialNets}. 
Multiple studies showed that domain alignment can be achieved by learning a domain classifier $C_{Disc}$ (the {\it discriminator}) with the parameters $\theta_D$ to distinguish between the feature vectors from source and target distributions and by using an adversarial loss to increase {\it domain confusion}~\cite{GaninJMLR16DomainAdversarialNN,TzengICCV15SimultaneousDeepTL,TzengCVPR17AdversarialDiscriminativeADDA}. The main, task-specific deep network (in our case, for SiS) then aims to learn a representation that fools the domain classifier, and therefore encourage encoders to produce domain-invariant features. Such features can then be used by the final classifier trained on the source data, to make predictions on the target data. Amongst the typical adversarial losses, we mention the min-max game proposed by Ganin~\etal~\cite{GaninJMLR16DomainAdversarialNN}:
\begin{align*}
  \Loss_{adv} = \min_{\theta_{F},\theta_{C}} \max_{\theta_{D}} \, &\{ \Exp_{\bfx_s \in \calX_\calS} [\mathcal{L}_{Task}(F(\bfx_s), \bfy_s)] \\
  &- \lambda \cdot \Exp_{\bfx \in \calX_\calS \cup \calX_\calT} [\mathcal{L}_{Disc}(F(\bfx), \bfy_d)] \},
\end{align*}
where $\mathcal{L}_{Task}$ is the loss associated with the task of interest (it depends on both the feature encoder parameters $\theta_F$ and the final classifier's parameters $\theta_C$), and $\mathcal{L}_{Disc}$ is a loss measuring how well a discriminator model parametrized by $\theta_D$ can distinguish whether features belong to source ($\bfy_d=1$) or target domain ($\bfy_d=0$). 
By alternatively training the discriminator $C_{Disc}$ to distinguish between domains and the feature encoder $F$
to ``fool'' it, one can learn domain agnostic features. Also training the encoder and the final classifier $C_{task}$ for the task of interest, guarantees that such features are not simply domain-invariant, but also discriminative.

An effective way to approach this min-max problem consists in introducing in the network a Gradient Reversal Layer (GRL)~\cite{GaninJMLR16DomainAdversarialNN} 
which reverses the gradient direction during the backward pass in backpropagation (in the forward pass, it is inactive). The GRL allows to train the discriminator and the encoder at the same time. 

A related but different approach by Tzeng~\etal~\cite{TzengCVPR17AdversarialDiscriminativeADDA} brings adversarial training for UDA closer to the original GAN formulation~\cite{GoodfellowNIPS14GenerativeAdversarialNets}. It splits the training procedure into two different phases: a fully discriminative one, where a module is trained on source samples, and a fully generative one, where a GAN loss is used to learn features for the target domain that mimic the source ones --- or, more formally, that are projected into the same feature space, on which the original classifier is learned. This second step can be carried out by approaching the following min-max game
\begin{align*}
  \Loss_{GAN} = \min_{\theta_{F_\calT},\theta_{F_S}} \max_{\theta_{D}} \, \left\{ \Exp_{\bfx_s \in \calX_\calS} [\log(C_{Disc}(F_S(\bfx_s)))] \right.
  \\
    +  \left. \Exp_{\bfx_t \in \calX_\calT} [\log(1-C_{Disc}(F_T(\bfx_t)))] \right\},
\end{align*}
\noindent
where $C_{Disc}$ is the discriminator, and both $(F_S$ and $(F_T$ are initialized with the weights pre-trained by supervised learning on the source data.

\begin{table*}
\begin{center}
{
\footnotesize
\begin{NiceTabular}{|l|c|c|c|c|c|c|c|c|}[code-before = 
\rowcolor{gray!10}{4}
\rowcolor{gray!10}{6}
\rowcolor{gray!10}{8}
\rowcolor{gray!10}{10}
\rowcolor{gray!10}{12}
\rowcolor{gray!10}{14}
\rowcolor{gray!10}{16}
\rowcolor{gray!10}{18}
\rowcolor{gray!10}{20}
\rowcolor{gray!10}{22}
\rowcolor{gray!10}{24}
\rowcolor{gray!10}{26}
\rowcolor{gray!10}{28}
\rowcolor{gray!10}{30}
\rowcolor{gray!10}{32}
\rowcolor{gray!10}{34}
\rowcolor{gray!10}{36}
\rowcolor{gray!10}{38}
\rowcolor{gray!10}{40}
\rowcolor{gray!10}{42}
\rowcolor{gray!10}{44}
\rowcolor{gray!10}{46}
\rowcolor{gray!10}{48}
\rowcolor{gray!10}{50}
]
\toprule

\textbf{Citation} & \textbf{Segmentation} & \textbf{ Image} 
  & \textbf{Network} &  \textbf{Sha-} 
    & \textbf{CW}  & \textbf{Output} & \textbf{Comple-} & \textbf{Specificity}\\
  & \textbf{Network} & \textbf{Level} & \textbf{Level} 
         & \textbf{red} & \textbf{feat.} & \textbf{Level}  & \textbf{mentary} & \\
\midrule

Hoffman \etal'16~\cite{HoffmanX16FCNsInTheWildPixelLevelAdversarialDA} & FCNs & - &  mDC   & \checkmark & \checkmark & mInstLoss & - &  class-size hist transfer\\
%
Chen \etal'17~\cite{ChenICCV17NoMoreDiscriminationCrossCityAdaptationRoadSceneSegmenters} & dFCN & - & DC &  \checkmark & \checkmark &- & SelfT & static obj prior \\
Perone \etal'17~\cite{PeroneNI19UDAMedicalImSegmSelfEnsembling} &  UNet & Aug & EMA &  - & - & SemCons & SelfEns & Dice loss/medical \\
%
Chen \etal'18~\cite{ChenCVPR18RoadRealityOrientedDASemSegmUrbanScenes} & DLab/PSPN  &  - & DC  & - & - & - & Distill & spatial aware adaptation \\ 
Huang \etal'18~\cite{HuangECCV18DomainTransferThroughDeepActivationMatching} & ENet &- & DC &  - & - & - & - & Jensen-Shannon divergence\\
Hong \etal'18~\cite{HongCVPR18ConditionalGANStructuredDA} & FCNs & - & DC &  \checkmark & \checkmark & - & - &  CGAN/target-like features \\
Hoffman \etal'18~\cite{HoffmanICML18CyCADACycleConsistentAdversarialDA} & FCN & S$\leftrightarrow$T & DC  & - &- & SemCons  &   - &  CycleGAN \\
Li \etal~'18\cite{LiBMVC18SemanticAwareGradGANVirtualToRealUrbanSceneAdaption}& UNet& S$\leftrightarrow$T & - & - & \checkmark & - & - & PatchGAN/semantic-aware gradient \\
Murez \etal'18~\cite{MurezCVPR18ImageToImageTranslationDA} & dFCN  & S$\leftrightarrow$T & DC    &- & - & SemCons &- & dilated DenseNet\\
Saito \etal'18~\cite{SaitoCVPR18MaximumClassifierDiscrepancyUDA}   & FCN & - &  - &  \checkmark & - & MCD &  CoT  &  minimize/maximize  discrepancy\\
Saito \etal'18~\cite{SaitoICLR18AdversarialDropoutRegularization} & FCN &  - & -   &  \checkmark & - & MCD &  CoT & adversarial drop-out  \\
%
Tsai \etal'18~\cite{TsaiCVPR18LearningToAdaptStructuredOutputSpaceSemSegm} & DeepLab & - & -  & \checkmark & - &  mDC  &  - & multi-level predictions\\ 
Wu \etal~'18\cite{WuECCV18DCANDualChannelWiseAlignmentNetworksUDA} & FCN &  S$\rightarrow$T  & DM &  \checkmark & - & -  & -  & channel-wise Gramm-matrix align \\
Zhu \etal'18~\cite{ZhuECCV18PenalizingTopPerformersConservativeLossSemSegmDA} & FCN &  S$\leftrightarrow$T  & - & \checkmark & - &  -& - & conservative loss\\
Zou \etal'18~\cite{ZouECCV18UnsupervisedDASemSegmClassBalancedSelfTraining}&   FCN &  -  & - & \checkmark &  - & -  & SelfT & self-paced curriculum/spacial priors\\
%
Chang \etal'19~\cite{ChangCVPR19AllAboutStructureDASemSeg} & DeepLab & S$\leftrightarrow$T & -   & - &  - &   DC &- & domain-specific encoders/percL  \\
Chen \etal'19~\cite{ChenCVPR19CrDoCoPixelLevelDomainTransferCrossDomainConsistency} & FCN/DRN  & S$\leftrightarrow$T & DC & - &  - &   SemCons & -  & KL cross-domain consistency \\ 
Chen \etal'19~\cite{ChenICCV19DASemSegmentationMaximumSquaresLoss} &  DeepLab & - & - &  \checkmark & - & - & SelfT & max squares loss/img-wise weights \\
Choi \etal'19~\cite{ChoiICCV19SelfEnsemblingGANBasedDataAugmentationDASemSegm} &ASPP  & S$\rightarrow$T  & EMA & - & - & SemCons & SelfEns &  target-guided+cycle-free augm. \\
Du \etal'19~\cite{DuICCV19SSFDANSeparatedSemanticFeatureDASemSegm}
& FCN &  -& DC &  \checkmark & \checkmark & - & PL  &class-wise adversarial reweighting \\
Lee \etal'19~\cite{LeeCVPR19SlicedWassersteinDiscrepancyUDA} &  PSPNet & - & - & \checkmark  & -  & MCD & CoT &   sliced Wasserstein discrepancy \\
Luo \etal'19~\cite{LuoCVPR19TakingACloserLookDomainShiftSemanticsConsistentDA} & FCN/DLab & - & -  &   \checkmark   &   - &  MCD & CoT &  local consistency/self-adaptive weight  \\
Li \etal'19~\cite{LiCVPR19BidirectionalLearningDASemSegm} & DeepLab & S$\leftrightarrow$T &  -  & \checkmark &  -   &  DC & SelfT  & perceptual loss (percL) \\ 
Lian \etal'19~\cite{LianICCV19ConstructingSelfMotivatedPyramidCurriculumsCrossDomainSemSegm} & FCN/PSPN  & - & - &  \checkmark  & -  & - & CurrL & self-motivated pyramid curriculum \\
Luo \etal'19~\cite{LuoICCV19SignificanceAwareInformationBottleneckDASemSegm} & FCN/DLab & - & DC  &   \checkmark & - & - & - & signif.-aware information bottleneck\\
Shen \etal'19~\cite{ShenX19RegularizingProxiesMultiAdversarialTrainingUDASS} & ASPP & - & mDC &   \checkmark & \checkmark  & ConfMap  & SelfT  &   cls+adv-conf./class-balance weighs  \\
Xu \etal'19~\cite{XuAAAI19SelfEnsemblingAttentionNetworksSemSegm} & DeepLab & Aug & EMA &  - & - & SemCons & SelfEns & 
self-ensembling attention maps \\
Vu \etal'19~\cite{VuCVPR19ADVENTAdversarialEntropyMinimizationDASemSegm} & DLab/ASPP & - & - &\checkmark &  - &  DC  & TEM &  entropy map align./class-ratio priors\\
Huang \etal'20~\cite{HuangECCV20ContextualRelationConsistentDASemSegm} &  DeepLab & - & - & \checkmark & \checkmark &  DC & TEM & local contextual-relation \\
%
 Lv \etal~\cite{LvCVPR20CrossDomainSemSegmentationInteractiveRelationTransfer} & FCN/DLab &  - &  - & \checkmark  & - & SemCons & SelfT & course-to-fine segm. interaction \\
Musto \etal'20~\cite{MustoBMVC20SemanticallyAdaptiveI2ITranslationDASemSegm} & FCN/DLab & S$\leftrightarrow$T & -  & \checkmark & - &  SemCons&  SelfT & spatially-adaptive normalization  \\
Pan \etal'20~\cite{PanCVPR20UnsupervisedIntraDASemSegmSelfSupervision}& DeepLab & - & -  &   - &  - & DC & CurrL & align entropy maps/easy-hard split \\
Toldo \etal'20~\cite{ToldoIVC20UDAMobileSemSegmCycleConsistencyFeatureAlignment} & DeepLab & S$\leftrightarrow$T & DC  & \checkmark & - & SemCons &    - &  MobileNet \\
Wang \etal'20~\cite{WangCVPR20DifferentialTreatmentStuffThingsDASemSegm} & ASPP & - & DC &  \checkmark &  \checkmark & - & SelfT &
disentangled things and stuff \\
Yang \etal'20~\cite{YangCVPR20PhaseConsistentEcologicalDA}& Unet  &S$\leftrightarrow$T  & - &  \checkmark & - & -  &  SelfT & phase cons./cond. prior network\\
Yang \etal'20~\cite{YangCVPR20FDAFourierDASemSegm} & FCN/DL &  S$\rightarrow$T & - &  \checkmark & -   & - & SelfT & Fourier transform/low-freq. swap \\
Yang \etal'20\cite{YangAAAI20AnAdversarialPerturbationOrientedDASS} & DeepLab  & - & advF  & \checkmark & - & 
DC & TEM & adv. attack/feature perturbation\\ %
Yang \etal'20~\cite{YangECCV20LabelDrivenReconstructionDASemSegm}& FCN/DLab & T$\rightarrow$S & - &  \checkmark   & - & DC & SelfT & reconstruction from predictions\\
Zang \etal'20~\cite{ZhangCVPR20TransferringRegularizingPredictionSemSegm} &  FCN &  - & DC &  \checkmark & - & LocCons & - &
patch+cluster+spatial consistency\\
Zheng \etal'20~\cite{ZhengIJCAI20UnsupervisedSceneAdaptationwithMemoryRegularizationInVivo}& PSPNet & - & -  & \checkmark & - & MCD  & CoT &  memory regularization (KL) \\
Araslanov \etal'21~\cite{AraslanovCVPR21SelfSupervisedAugmentationConsistencyAdaptingSemSegm}& DeepLab & Aug & EMA & - &  & SemCons  & SelfT & self-sup/imp. sampling/focal loss\\
Cheng \etal'21~\cite{ChengICCV21DualPathLearning4DASS} &
DLab/FCN &S$\leftrightarrow$T &   - & - & - & SemCons & SelfT & dual perceptual loss/dual path DASS\\
Guo \etal'21~\cite{GuoCVPR21MetaCorrectionDomainAwareMetaLossCorrectionDASS}  & DeepLab  & - & - &  \checkmark & - & - & SelfT &  meta-learning/meta-loss correction \\
Toldo~\etal'21~\cite{ToldoWACV21UDASemSegmOrthogonalClusteredEmbeddings}   &  DeepLab & - & Clust. &  \checkmark &  \checkmark & &  TEM & discriminative clustering \\
Truong \etal'21~\cite{TruongICCV21BiMaLBijectiveMaximumLikelihoodDASS}  & DeepLab & - & -&  \checkmark & - &  - & TEM & bij. max. likelihood/local consistency
\\
%
Wang \etal'21~\cite{WangICCV21UncertaintyAwarePseudoLabelRefineryDASS} & DeepLab &  S$\rightarrow$T & -& \checkmark  & - & DC  & SelfT & target-guided uncertainty rectifying\\
Wang \etal'21~\cite{WangAAAI21ConsistencyRegularizationSourceGuidedPerturbationUDASemSegm} & FCN & S$\leftrightarrow$T & EMA & - & - & - & SelfEns  & AdaIn/class-balanced
reweighting \\
%
Yang \etal'21~\cite{YangICCV21ExploringRobustnessDASS} &
DeepLab & S$\rightarrow$T & - & \checkmark  & - & contrL & SelfT &
adv. attack/adv. self-supervised loss \\  

\bottomrule
\end{NiceTabular}
} 
\end{center}
\caption{Summary of the state-of-the-art methods, schematized according to their characteristics. 
\textbf{Segmentation Network}: The neural network used as a backbone, 
\textbf{Image Level}: alignment at image level (by using style transfer), from source to target S$\rightarrow$T, target to source S$\leftarrow$T or both  S$\leftrightarrow$T; Aug stands for specific data augmentation strategies. 
\textbf{Network Level}: alignment at feature level 
where DC stands for adversarial domain confusion at single or  multiple (mDC) feature levels,  DM is discrepancy minimization between feature distributions, Clust stands for feature level clustering, AdvF for adversarial features.  Alternatively, EMA (exponential moving average) is model parameter adaptation
from  student to a teacher model or inversely.
\textbf{Shared}: the  parameters of the segmentation network are shared (\checkmark ) or  at least partially domain specific (-). 
\textbf{CW}:  class-wise feature alignment or clustering.
\textbf{Output Level}: alignment 
or regularization at output level: DC/DM  with  features  extracted from the predicted maps
or confusion maps (ConfMap),
MCD  stands for adversarial training of multi-classifier discrepancy. Further notations:  contrL (contrastive loss),
SemCons (semantic consistency loss between  predicted segmentation  maps), LocCons (local consistency),  mInstLoss (multi-instance loss).
\textbf{Complementary} indicates which complementary techniques are used, such as
exploiting pseudo labels (PL), 
self-training (SelfT), curriculum learning (CurrL), target (conditional) entropy minimisation (TEM),
CoT (co-training),
self-ensembling (SelfEns), 
model distillation (Distill).
\textbf{Specificity} reports important elements that are not covered by the previously described common characteristics.
}
\label{tab:semsegm_methods}
\end{table*}


\subsection{Domain alignment}
\label{sec:dass_siamese}

\begin{figure}[ttt]
\begin{center}
\includegraphics[width=0.47\textwidth]{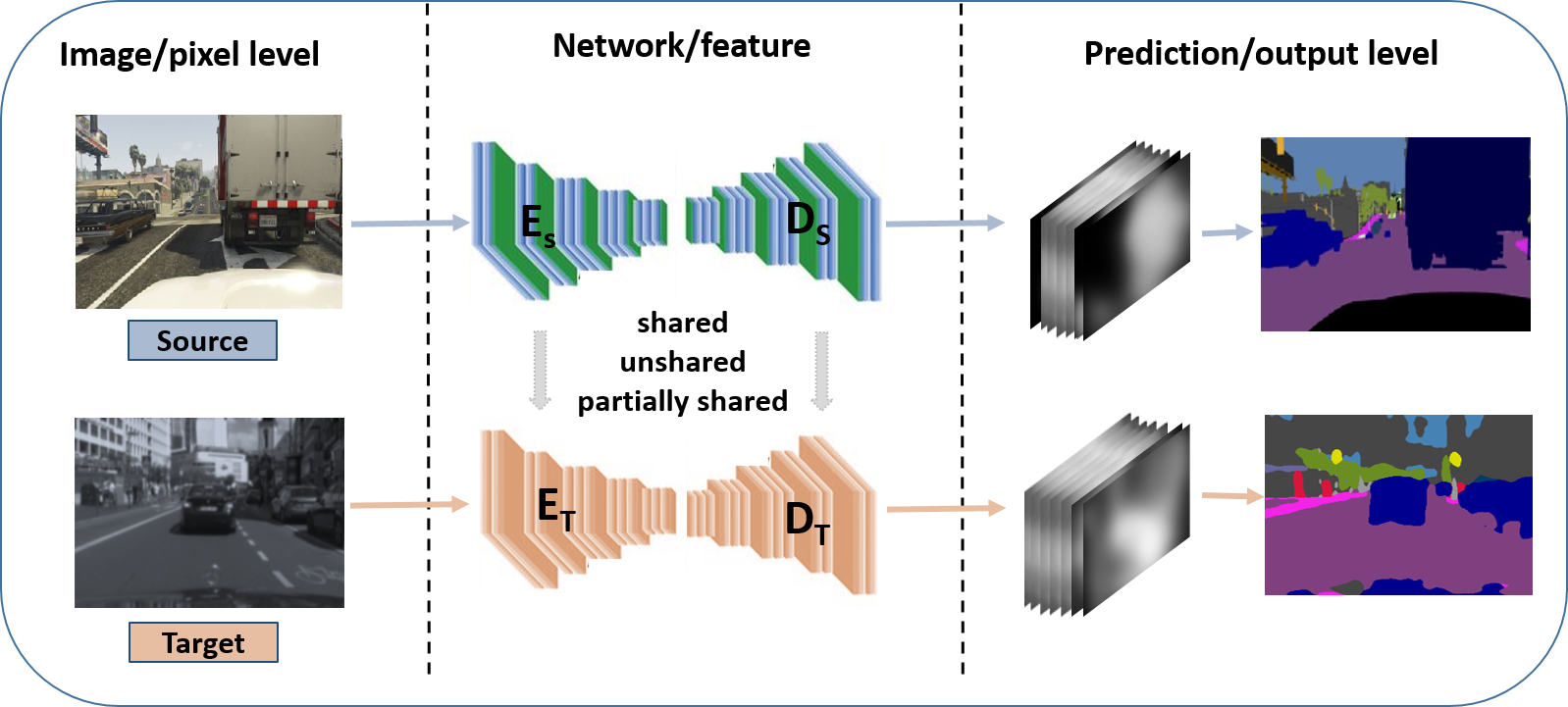}
\caption{Several DASiS models adopt a Siamese architecture with one branch per domain, where the alignment is carried out at different levels of the pipeline: at {\it pixel level}, by transferring the image style from one domain to another; at {\it network level}, by aligning the image features, derived from activation layers and sharing (full or partial) the network parameters; and finally at the {\it output level}, by aligning the class prediction maps.}
\label{fig:AdaptLevels}
\end{center}
\end{figure}

\myparagraph{Preliminaries}
Since the advent of representation learning solutions in most of machine learning applications, UDA research has witnessed a shift towards end-to-end solutions to learn models that may perform well on target samples. In image classification, a very successful idea has been to learn a representation where the source and target samples results {\it aligned} --- that is, the source and target distributions are close in the feature space under some statistical metrics.

\begin{figure*}[ttt]
\begin{center}
\includegraphics[width=0.9\textwidth]{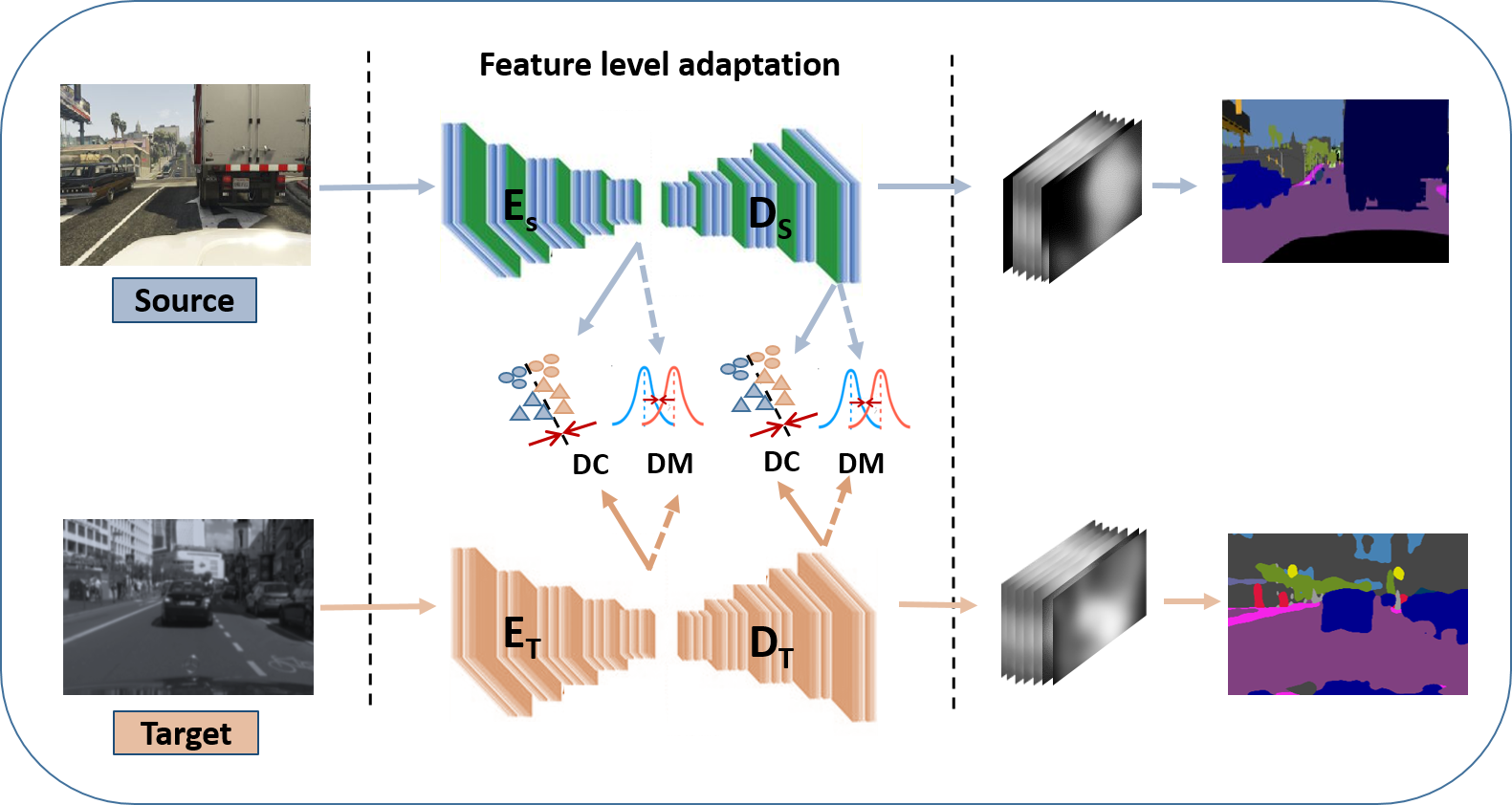}
\caption{In generic DA, domain alignment is often performed in a single latent representation space.
In DASiS, the alignment is often done at multiple layers, by discrepancy minimization between feature distributions or by adversarial learning relying on a domain classifier (DC) to increase domain confusion. 
Encoders and decoders of the segmentation network are often shared: $E_S=E_T$, $D_S=D_T$.}
\label{fig:FLAdapt}
\end{center}
\end{figure*}

This~\textit{alignment} is often achieved by means of a Siamese
architecture~\cite{Bromley93IJPRAISignatureVerificationTimeDelayNN} with two streams, 
each corresponding to a semantic segmentation model: one stream is aimed at processing source samples and the other at processing the target ones. The parameters of the two streams can be shared, partially shared or domain specific; generally, the backbone architectures of both streams are initialized with weights that are pre-trained on the source set (as shown in~\fig{AdaptLevels}). 
The Siamese network is typically trained
with a loss comprising two terms. For what concerns SiS, one term is the standard~\textit{pixel-wise cross-entropy} loss (referred in this paper also as $\Loss_{Task}$), measuring performance on source samples, for which the ground truth annotations are available: 
\begin{align*}
\Loss_{ce}  =  - \mathbb{E}_{
(\calX_\calS, \calY_\calS) }  
 \left[  \sum_{h,w,c}{\bfy_s^{(h,w,c)} \cdot  \log(p^{(h,w,c)}(F_S(\bfx_s))}) \right]
\end{align*}
where $p^{(h,w,c)}(F_S(\bfx_s))$ is a probability of class $c$ at pixel $\bfx_s^{(h,w)}$ and $\bfy_s^{(h,w,c)}$ is the pixel's true class.

The second term is a~\textit{domain alignment} loss that measures the distance between source and target samples. The alignment can be addressed at different levels of the pipeline as illustrated in~\fig{AdaptLevels}, namely, at pixel, network (feature) and output (prediction) levels. We will detail them in the following paragraphs.

While aligning the marginal feature distributions tends to reduce the domain gap, it can be sub-optimal as it does not explicitly take the specific task of interest (in this case, SiS) into account during the domain alignment~\cite{ZhaoICML19OnLearningInvariantReprDA}. To overcome these weaknesses, several works have been proposed to leverage the class predictions during the alignment, what we call output level alignment. Finally, the interest is growing for adaptation at pixel level. Indeed, the shift between two domains is often strongly related to visual appearance variations such as day \vs night, seasonal changes, synthetic \vs real. By exploiting the progress of image-to-image translation and style transfer brought by deep learning based techniques~\cite{HuangICCV17ArbitraryStyleTransferRealTimeAdaptiveInstanceNormalization,ZhuICCV17UnpairedI2ICycleConsistentAdversarialNetworks}, several DASiS methods have been proposed that explicitly account for such stylistic domain shifts by performing an alignment at image level. 

In the following paragraphs, we detail methods that propose different alignment solutions between source and target at various levels of the segmentation pipeline. Note that --- as shown in Table~\ref{tab:semsegm_methods} --- many approaches apply alignment at multiple levels.

\begin{figure*}[ttt]
\begin{center}
\includegraphics[width=0.95\textwidth]{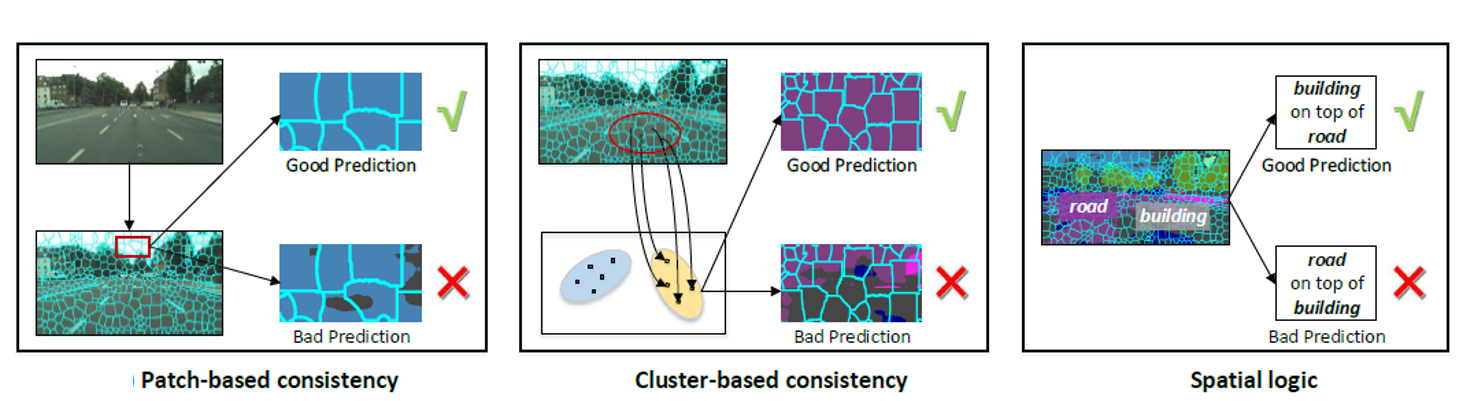}
\caption{To improve the domain alignment, Zhang~\etal~\cite{ZhangCVPR20TransferringRegularizingPredictionSemSegm} propose to reduce patch-level, cluster-level and context-level inconsistencies.
(Image courtesy of~\cite{ZhangCVPR20TransferringRegularizingPredictionSemSegm}).}
\label{fig:ZhangRegularize}
\end{center}
\end{figure*}

\subsubsection{Feature-level adaptation} 
\label{sssec:image-level}

Generic DA solutions proposed for image classification perform domain alignment in a latent space by minimizing some distance metrics, such as the maximum mean discrepancy (MMD)~\cite{LongICML15LearningTransferableFeaturesDAN} between feature distributions of source and target data, or by adversarially training a domain discriminator to increase {\it domain confusion}~\cite{GaninJMLR16DomainAdversarialNN,TzengICCV15SimultaneousDeepTL,TzengCVPR17AdversarialDiscriminativeADDA}. 
Both approaches scale up to semantic segmentation problems --- in particular, adversarial training has been largely and successfully applied and combined with other techniques.
 
In DASiS, we consider more complex models to tackle the SiS task.
We recall that adaptation i SiS is more challenging than in image classification, due to the structural complexity and the scale factor 
of the task which is rather difficult to fully capture and handle by simple alignment of the latent representations (activation layers) between domains.
Therefore, the domain alignment is often carried out at different layers of the network, by minimizing the feature distribution discrepancy~\cite{BermudezChaconISBI18ADomainAdaptiveUNetElectronMicroscopy} or by adversarial training via a domain classifier to increase domain confusion~\cite{HoffmanX16FCNsInTheWildPixelLevelAdversarialDA,HuangECCV18DomainTransferThroughDeepActivationMatching,ShenX19RegularizingProxiesMultiAdversarialTrainingUDASS} (see Figure~\ref{fig:FLAdapt}).
While some works consider simply a global representation of the image (by flattening or pooling the activation map)~\cite{HuangECCV18DomainTransferThroughDeepActivationMatching}, most often pixel or grid-wise~\cite{HoffmanX16FCNsInTheWildPixelLevelAdversarialDA,ChenICCV17NoMoreDiscriminationCrossCityAdaptationRoadSceneSegmenters} or region-wise (super-pixel) representations ~\cite{ZhangCVPR20TransferringRegularizingPredictionSemSegm} are used. 
Furthermore,  to improve the model performance on the target data such methods are often combined with some prior knowledge or specific tools as discussed below and in section \ref{sec:dass_improvements}.

In their seminal work, Hoffman~\etal~\cite{HoffmanX16FCNsInTheWildPixelLevelAdversarialDA}, combine the distribution alignment with the class-aware constrained multiple instance loss used to transfer the spatial layout. 
Chen~\etal~\cite{ChenICCV17NoMoreDiscriminationCrossCityAdaptationRoadSceneSegmenters} consider global and class-wise domain alignment and address it via adversarial training. In particular, they rely on local class-wise domain predictions over image grids assuming that the composition/proportion of object classes across domains (\ie, different urban environments) is similar.
Hong~\etal~\cite{HongCVPR18ConditionalGANStructuredDA} rely on a conditional generator that transforms the source features into target-like features, using a multi-layer perceptron as domain discriminator. Assuming that the decoding these target-like feature maps preserve the semantics, they are used with the corresponding source labels within an additional cross-entropy loss to make the model more suitable  for the target data.

Zhang~\etal~\cite{ZhangCVPR20TransferringRegularizingPredictionSemSegm}, to improve alignment, explore three label-free constraints as model regularizer, by enforcing patch-level, cluster-level and context-level semantic prediction consistencies at different levels of image formation (see~\fig{ZhangRegularize}).
Lv~\etal~\cite{LvCVPR20CrossDomainSemSegmentationInteractiveRelationTransfer}
propose a so called Pivot Interaction Transfer, which consists in a semantic consistency loss between image-level and  pixel-level semantic information. This is achieved by training the model with both a fine-grained component producing pixel-level segmentation and coarse-grained components generating class activation maps obtained by multiple region expansion units trained with image-level category information independently.

\begin{figure*}[ttt]
\begin{center}
\includegraphics[width=0.9\textwidth]{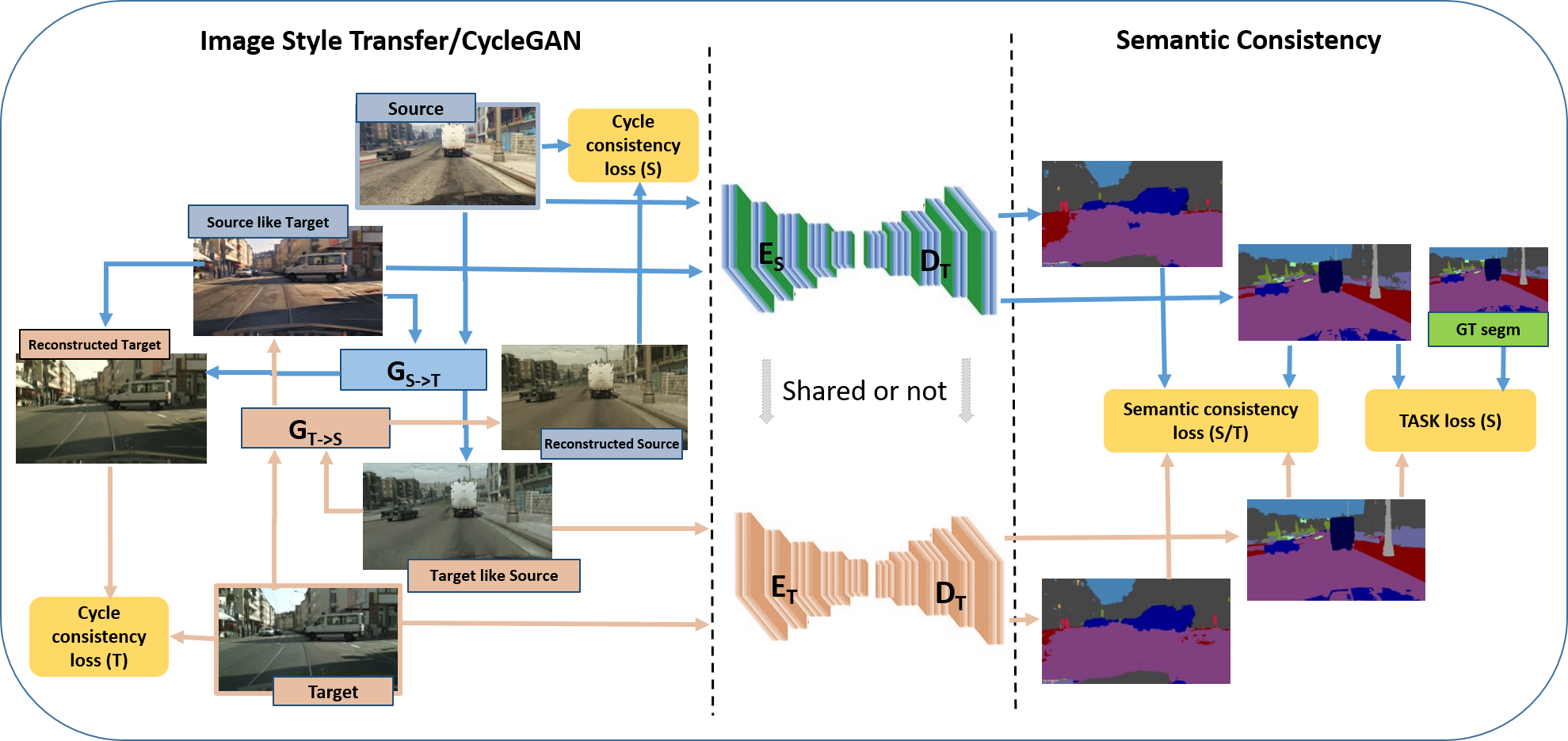}
\caption{In image-level adaptation that relies on image style transfer (IST), the main idea is to translate the “style” of the target domain to the source data and/or the source style to the target domain.
In order to improve the style transfer, often the style transferred image is translated back to the original domain allowing to use a cyclic consistency reconstruction loss. The “style transferred” source images inherit the semantic content of the source and hence its pixel-level labeling, that allows the segmentation network to learn a model suitable for the target domain. On the other hand, the target and source-like target image share the content and therefore imposing that their predicted segmentation should match (semantic consistency loss) acts as a regularization and helps improving the model performance in the target domain.}
\label{fig:IST}
\end{center}
\end{figure*}

\subsubsection{Image-level adaptation} 
\label{sssec:image-level}

This class of methods relies on image style transfer (IST) methods, where the main idea is to transfer the domain ``style'' (appearance)
from target to source, from source to target, or consider both
(see illustration in~\fig{IST}).
The ``style transferred'' source images maintain the semantic content of the source, and hence its pixel-level labeling too, while their appearance results more similar to the target style --- helping the network to learn a model more suitable for the target domain.

Image-to-image translation for UDA has been pioneered within the context of image classification~\cite{BousmalisCVPR17UnsupervisedPixelLevelDAwGANs,LiuNIPS16CoupledGANs,TaigmanICLR17UnsupervisedCrossDomainImageGeneration}; typically, such methods employ Generative Adversarial Networks (GANs) \cite{GoodfellowNIPS14GenerativeAdversarialNets} to transfer the target images' style into one that resembles the source style. This approach has been proved to be a prominent strategy also 
within DASiS~\cite{ChangCVPR19AllAboutStructureDASemSeg,ChenCVPR19CrDoCoPixelLevelDomainTransferCrossDomainConsistency,HoffmanICML18CyCADACycleConsistentAdversarialDA,MurezCVPR18ImageToImageTranslationDA,ToldoIVC20UDAMobileSemSegmCycleConsistencyFeatureAlignment,SankaranarayananCVPR18LearningSyntheticDataDomainShiftSemSegm,WuECCV18DCANDualChannelWiseAlignmentNetworksUDA,YangCVPR20PhaseConsistentEcologicalDA}. Still, as in the case of feature alignment, for a better adaptation most methods combine the image translation  with other ingredients (see also Table~\ref{tab:semsegm_methods}), most often with self-training and different consistency regularization terms (detailed in section \ref{sec:dass_improvements}). 

The most used regularization terms are the cycle consistency loss and the semantic consistency loss proposed by Hoffman~\etal~\cite{HoffmanICML18CyCADACycleConsistentAdversarialDA}, who proposed CyCADA, one of the first method that adopted image-to-image translation --- and in particular the consistency losses pioneered by Cycle-GAN~\cite{ZhuICCV17UnpairedI2ICycleConsistentAdversarialNetworks} --- for the DASiS problem.
The cycle consistency loss is defined as follows
\begin{align*}
	\begin{split}
	\Loss_{cycle} &  =
	 \Exp_{\bfx_s \sim \calX_\calS} \big[ \| G_{T \rightarrow S} \left( G_{S \rightarrow T} \left(\bfx_s\right) \right) - \bfx_s \|_k\big] \\
	& + \Exp_{\bfx_T\sim \calX_\calT}  \big[ \| G_{S \rightarrow T} \left( G_{T \rightarrow S} \left(\bfx_t\right) \right) - \bfx_t \|_k\big].
	\end{split}
\end{align*}
where $G_{S \rightarrow T}$ and $G_{T \rightarrow S}$ are 
the image generators  that learn to map the "style" 
from source to target and target to source, respectively 
and $\|\cdot \|_k$ is the  L$k$ loss, where most often the L1  or the L2
loss is used. 
In short, this loss encourages the preservation of structural properties during the style transfer, while the semantic consistency loss
\begin{align*}
	\begin{split}
  \Loss_{SemCons} & = 
  \mathcal{L}_{Task} \left( F_S(G_{S \rightarrow T}(\bfx_s)), p(F_S(\bfx_s)) \right) \\ 
& + \mathcal{L}_{Task} \left( F_\calS(G_{T \rightarrow S}(\bfx_t)), p(F_S(\bfx_t))  \right),
	\end{split}	
\end{align*}
enforces  an image to be labeled identically before and after translation. The task loss  $\mathcal{L}_{Task}$ here is the source pixel-wise cross-entropy, but instead of using GT label maps, it is 
used with the pseudo-labeled predicted  maps $p(F_S(\bfx_s)) =\argmax (F_S(\bfx_s))$ respectively $p(F_S(\bfx_t)) =\argmax (F_S(\bfx_t))$.

Inspired by CyCADA, several approaches tried to refine IST for the DASiS problem. Murez~\etal~\cite{MurezCVPR18ImageToImageTranslationDA} propose a method that simultaneously learns domain specific reconstruction with cycle consistency and domain agnostic feature extraction, then it learns to predict the segmentation from these agnostic features. In the IST based method proposed in~\cite{ZhuECCV18PenalizingTopPerformersConservativeLossSemSegmDA} the classical cross-entropy loss is replaced by a so-called ``Conservative Loss'' that penalizes the extreme cases (that is, for which performance is very good or very bad) enabling the network to find an equilibrium between its discriminative power and its and domain-invariance.

Toldo~\etal~\cite{ToldoIVC20UDAMobileSemSegmCycleConsistencyFeatureAlignment} perform image-level domain adaptation with Cycle-GAN~\cite{ZhuICCV17UnpairedI2ICycleConsistentAdversarialNetworks} and feature-level adaptation with a consistency loss between the semantic maps. Furthermore, they consider as backbone a lightweight MobileNet-v2 architecture which allows the model deployment on devices with limited computational resources such as the ones used in autonomous vehicles.

Li~\etal~\cite{LiBMVC18SemanticAwareGradGANVirtualToRealUrbanSceneAdaption} propose a semantic-aware Grad-GAN that aims at transferring personalized styles for distinct semantic regions. This is achieved by a soft gradient-sensitive objective for keeping semantic boundaries and a semantic-aware discriminator for validating the fidelity of personalized adaptions with respect to each semantic region.

The method introduced by Wu~\etal~\cite{WuECCV18DCANDualChannelWiseAlignmentNetworksUDA} jointly synthesizes images and performs segmentation by fusing channel-wise distribution alignment to preserve the spatial structure with semantic information in both the image generator and the segmentation network. In particular the generator synthesizes new images on-the-fly to appear target-like and the segmentation network refines the high level features before predicting semantic maps by leveraging feature statistics of sampled images from the target domain.

Chen~\etal~\cite{ChenCVPR19CrDoCoPixelLevelDomainTransferCrossDomainConsistency} rely on both image-level adversarial loss to learn image translation and feature-level adversarial loss to align feature distributions. Furthermore, they propose a bi-directional cross-domain consistency loss based on KL divergence, to provide additional supervisory signals for the network training, and show that this yields more accurate and consistent predictions in the target domain. 
  
Chang~\etal~\cite{ChangCVPR19AllAboutStructureDASemSeg} introduce the Domain Invariant Structure Extraction (DISE) method, which combines image translation with the encoder-decoder based image reconstruction where a set of shared and private encoders are used to disentangle high-level, domain-invariant structure information from domain-specific texture information. Domain adversarial losses and perceptual losses ensure 
the perceptual similarities between the translated images and their counterparts in the source or target domains and an adversarial loss in the output space ensures domain alignment and hence generalization to the target.

\begin{figure}[ttt]
\begin{center}
\includegraphics[width=.45\textwidth]{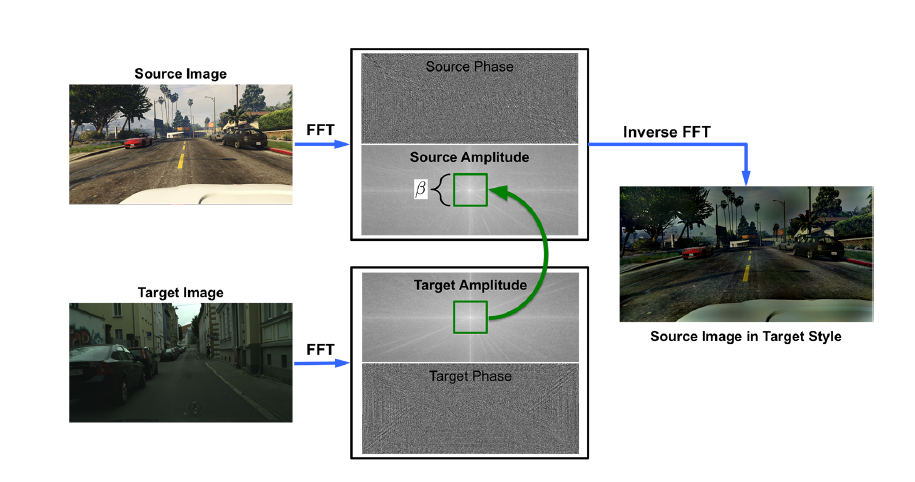}
\caption{Spectral transfer proposed by Yang and Soatto~\cite{YangCVPR20FDAFourierDASemSegm}, which relies on the Fourier Transform: the target ``style'' is mapped into that of the source images by swapping the low-frequency component of the spectrum of the images from the two domains.}
\label{fig:FDA}
\end{center}
\end{figure}

The approach by Li~\etal~\cite{LiCVPR19BidirectionalLearningDASemSegm} relies on bi-directional learning, proposing to move from a sequential pipeline --- where the SiS model benefits from the image-to-image translation network --- to a closed loop, where the two modules help each other. Essentially, the idea is to propagate information from semantic segmentation back to the image transformation network as a semantic consistent regularization --- and not limiting to propagate information from the image-to-image translation network to the segmentation module.

Cheng~\etal~\cite{ChengICCV21DualPathLearning4DASS} consider two image translation and segmentation pipelines from opposite domains to alleviate visual inconsistencies raised by image translation and  promote each other in an interactive manner. The source path assists the target path to learn precise supervision from source data, while the target path guides the source path to generate high quality pseudo-labels for self-training the target segmentation network.

Musto~\etal~\cite{MustoBMVC20SemanticallyAdaptiveI2ITranslationDASemSegm} propose a translation model from source to target, guided by semantic map of the source image using Spatially-Adaptive (de)normalization (SPADE) layers proposed by
Park~\etal~\cite{ParkCVPR19SemanticImageSynthesiswithSpatiallyAdaptiveNormalization}, followed by Instance Normalization layers~\cite{UlyanovX16InstanceNormalizationFastStylization}. Yang~\etal~\cite{YangECCV20LabelDrivenReconstructionDASemSegm} introduce a reconstruction network that relies on conditional GANs, which learns to reconstruct the source or the source-like target image from their respective predicted semantic label map. Finally, a perceptual loss and a discriminator feature matching loss are used to enforce the semantic consistency between the reconstructed and the original image features.

Finally, recent works propose solutions that do not rely primarily on GANs for image translation. For instance, the innovative approach by~\cite{YangCVPR20FDAFourierDASemSegm} relies on the Fourier Transform and its inverse to map the target ``style'' into that of the source images, by swapping the low-frequency component of the spectrum of the images from the two domains (see~\fig{FDA}). The same research team proposes to exploit the phase of the Fourier transform within a consistency loss~\cite{YangCVPR20PhaseConsistentEcologicalDA}; this guarantees to have an image-to-image translation network that preserves semantics.



\begin{figure}[ttt]
\begin{center}
\includegraphics[width=.45\textwidth]{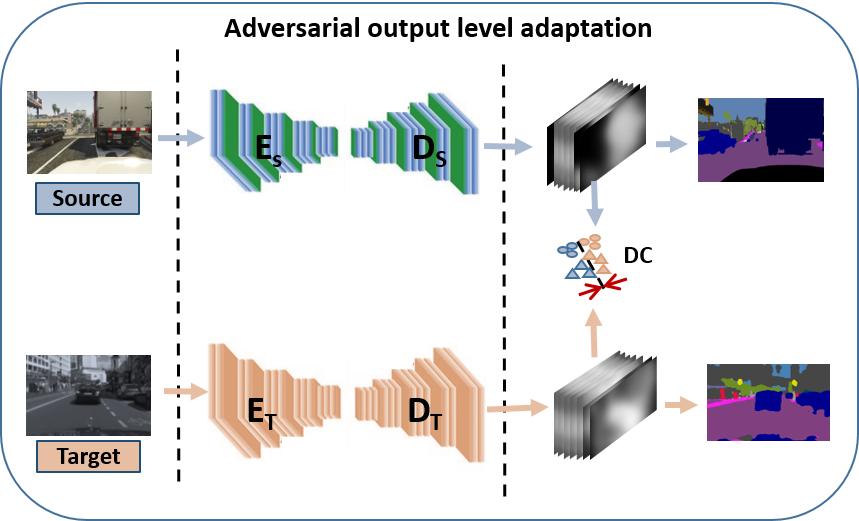}
\caption{Adversarial adaptation on the label prediction output space, where pixel-level representations are derived from the class-likelihood map and used to train the domain classifier.}
\label{fig:OutputLevelDA}
\end{center}
\end{figure}

\subsubsection{Output-level adaptation}  
\label{sssec:output-level}

To avoid the complexity of high-dimensional feature space adaptation, several papers proposed to perform instead adversarial adaptation on the low-dimensional label prediction output space, defined by the class-likelihood maps. In this case,  pixel-level representations are derived from these maps and then similarly to the approaches described in Section~\ref{sssec:image-level}, 
domain confusion can be applied in the derived ``feature space'' (see left image in~\fig{OutputLevelDA}).
 
Adversarial learning in the output space was first proposed by Tsai~\etal~\cite{TsaiICCV19DAStructuredOutputDiscriminativePatchRepresentations} where they 
learn a discriminator to distinguish whether the segmentation predictions were from the source or from the target domain. To make the model adaptation more efficient, auxiliary pixel-level semantic and domain classifiers are added at multiple layer's of the network and trained jointly.

Vu~\etal~\cite{VuCVPR19ADVENTAdversarialEntropyMinimizationDASemSegm} first derive so called ``weighted self-information maps'' $I^{(h,w,c)}_x=-p^{(h,w,c)} \log p^{(h,w,c)}$ and perform adversarial adaptation on the features derived from these maps.
Furthermore, showing that minimizing the sum of these weighted self-information maps is equivalent to direct entropy minimization, they jointly train the model with these two complementary entropy-based losses. 

Similarly, Pan~\etal~\cite{PanCVPR20UnsupervisedIntraDASemSegmSelfSupervision} train a domain classifier on the entropy maps 
$E^{(h,w)}_x=-\sum_c I^{(h,w,c)}_x$ 
to reduce the distribution shift between the source and target data. To further improve model performance in the target domain, they separate the target data into easy and hard samples --- relying on the entropy --- and try to diminish the gap between those predictions by so called intra-domain adversarial training on the corresponding entropy maps. 

Output level adversarial learning has often been used in combination with
image-level style transfer and self-training ~\cite{ChangCVPR19AllAboutStructureDASemSeg,LiCVPR19BidirectionalLearningDASemSegm,WangICCV21UncertaintyAwarePseudoLabelRefineryDASS} (see Table~\ref{tab:semsegm_methods}). 

 

\subsection{Complementary techniques}
\label{sec:dass_improvements}

In the previous section, we mainly focused on
the core issue 
of domain alignment. In this section, we discuss other methodologies that can be coupled with several DASiS methods previously presented, generally focusing not explicitly on the domain alignment, but rather
on improving the segmentation model accuracy on the target data. As a part of the transfer learning, DASiS --- and UDA in general --- possess characteristics (domain separation, unlabeled target instances, etc.) that encourage researchers to integrate techniques from ensemble, semi- and self-supervised learning, often resulting in their mutual empowering. 
While being extensively used in UDA research, the methodologies detailed below are 
originated and hence 
applied in other branches of machine learning; for example,  {\it self-training with pseudo-labels}~\cite{LeeICMLW13PseudoLabelSimpleEfficientSSLMethodDNN} and {\it entropy minimization}~\cite{GrandvaletNIPS2004SSLEntropyMinimization} 
were originally devised for  semi-supervised learning; {\it curriculum learning} was originally devised as a stand-alone training paradigm~\cite{BengioICML2009CurriculumLearning}; {\it model distillation} and {\it self-ensembling} are recent deep learning techniques that allow training more accurate models.

\begin{figure}[ttt]
\begin{center}
\includegraphics[width=0.45\textwidth]{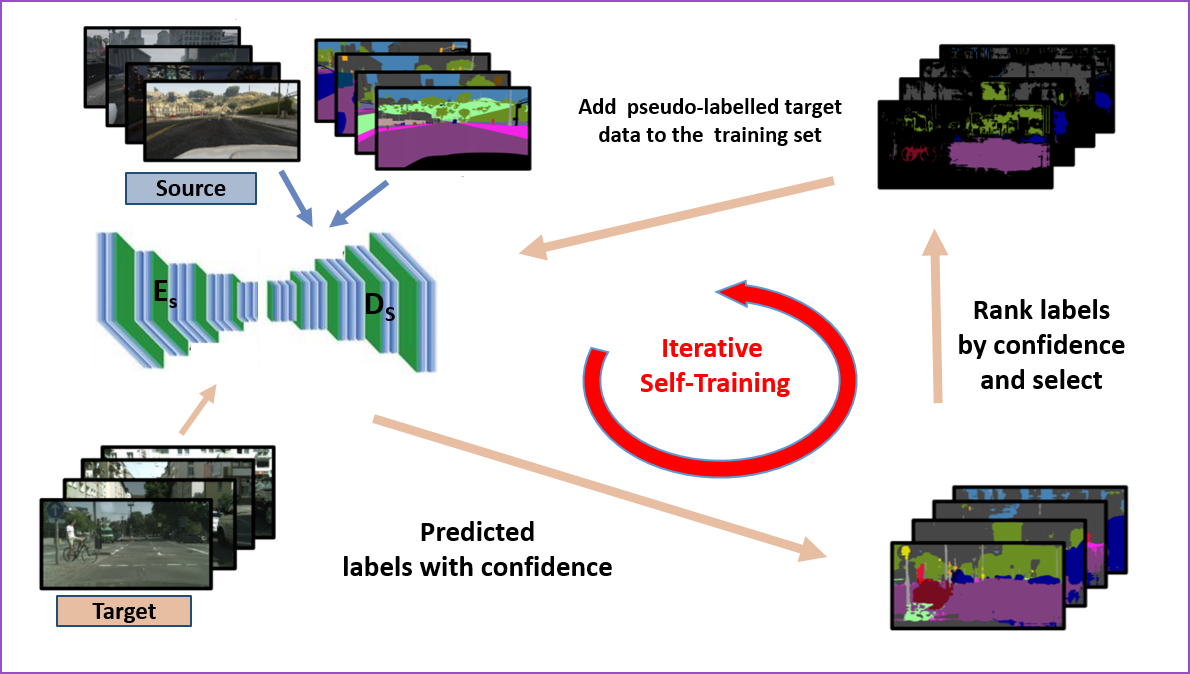}
\caption{Pseudo-labeling. Originated from semi-supervised learning, the idea is to generate pseudo-labels for the target data and to refine the model over iterations, by using the most confident labels from the target set.}
\label{fig:SelfTrainingSS}
\end{center}
\end{figure}

\myparagraph{Pseudo-labelling and self-training (SelfT)}
Originated from semi-supervised learning, the idea is to generate pseudo-labels for the target data and to iteratively refine (self-train) the model by using the most confident labels from the target set~\cite{LiCVPR19BidirectionalLearningDASemSegm,ZouECCV18UnsupervisedDASemSegmClassBalancedSelfTraining,KimCVPR20LearningTextureInvariantRepresentationDASemSegm,LiCVPR19BidirectionalLearningDASemSegm} (see illustration in~\fig{SelfTrainingSS}). Indeed, pseudo-labels are often error-prone, so it is important to select the most reliable ones and to progressively increase the set of pseudo-labels as the training progresses. To this end, different works \cite{LiECCV20ContentConsistentMatchingDASemSegm,WangCVPR20DifferentialTreatmentStuffThingsDASemSegm,ZouECCV18UnsupervisedDASemSegmClassBalancedSelfTraining} rely on the softmax of the model output as confidence measure. Pseudo-labels for which the prediction is above a certain threshold are assumed to be reliable; vice-versa, values below the threshold are not trusted. As shown in Table~\ref{tab:semsegm_methods}, self-training is one of the most popular complementary methods combined with domain alignment techniques. 

In the following, we list a few such methods with their particularities, aimed at further improving the effectiveness of self-learning. 
To avoid the gradual dominance of large classes on pseudo-label generation, 
Zou~\etal~\cite{ZouECCV18UnsupervisedDASemSegmClassBalancedSelfTraining} propose a class balanced self-training framework and introduce spatial priors to refine the generated labels. Chen~\etal~\cite{ChenICCV17NoMoreDiscriminationCrossCityAdaptationRoadSceneSegmenters} rely on static-object priors estimated for the city of interest by harvesting the time-machine of  Google Street View  to improve the soft pseudo-labels required by the proposed class-wise adversarial domain alignment.

Kim~\etal~\cite{KimCVPR20LearningTextureInvariantRepresentationDASemSegm} introduce a texture-invariant pre-training phase; in particular, the method relies on image-to-image translation to learn a better performing model at one stage, which is then adapted via the pseudo-labeling
in a second stage.

In order to regularize the training procedure, Zheng~\etal~\cite{ZhengIJCAI20UnsupervisedSceneAdaptationwithMemoryRegularizationInVivo}~average the predictions of two different sets of predictions from the same network. Relatedly, Shen~\etal~\cite{ShenX19RegularizingProxiesMultiAdversarialTrainingUDASS} combine the output of different discriminators with the confidence of a segmentation classifier, in order to increase the reliability of the pseudo-labels. To ``rectify'' the pseudo-labels, Zheng~\etal~\cite{ZhengIJCV21RectifyingPseudoLabelLearningUncertaintyEstimationDASS} propose to explicitly estimate the prediction uncertainty during training. They model the uncertainty via the prediction variance and involve the uncertainty into the optimization objective. 
 
Differently, Mei~\etal~\cite{MeiX20InstanceAdaptiveSelfTrainingUDA} propose an ``instance adaptive'' framework where pseudo-labels are generated via an ``adaptive selector'', namely a confidence-based selection strategy with a confidence threshold that is adaptively updated throughout training. Regularization techniques are also used to respectively smooth and sharpen the pseudo-labels and non-pseudo-labels regions.
 
In order to make self-training less sensitive to incorrect pseudo-labels,
Zou~\etal~\cite{ZouICCV19SConfidenceRegularizedSelfTraining} rely on soft pseudo-labels in the model regularization, forcing the network output to be smooth. Shin~\etal~\cite{ShinECCV20TwoPhasePseudoLabelDensificationDA} propose  a pseudo-label densification framework where a sliding window voting scheme is used to propagate confident neighbor predictions.  In a second phase  a confidence-based easy-hard classification selects images for self-training while a hard-to-easy adversarial learning pushes hard samples to be like easy ones. Tranheden~\etal~\cite{TranhedenWACV21DACSDACrossDomainMixedSampling} to mitigate low-quality pseudo-labels arising from the domain shift, propose to mix images from the two domains along with the corresponding labels and pseudo-labels and train the model while enforcing consistency between predictions of images in the target domain and images mixed across domains.

Zhang~\etal~\cite{ZhangNIPS19CategoryAnchorGuidedUDASS} propose a strategy where pseudo-labels are used both in a cross-entropy loss and in a ``category-wise distance loss'', where class-dependent centroids are used to assign pseudo-labels to training samples.


Li~\etal~\cite{LiECCV20ContentConsistentMatchingDASemSegm} propose to select source images that are most similar to the target ones  via ``semantic layout matching'' and ``pixel-wise similarity matching'' to select the pixels retained for the adaptation and used together with pseudo-labeled target samples to refine the model. Furthermore, entropy regularization is imposed on all the source and target images.

Guo~\etal~\cite{GuoCVPR21MetaCorrectionDomainAwareMetaLossCorrectionDASS} propose to improve the reliability of pseudo-labels via the  ``metaCorrection'' framework; they model the noise distribution of the pseudo-labels by introducing a ``noise transaction matrix'' that encodes inter-class noise transition relationship. The meta correction loss is further exploited to improve the pseudo-labels via a meta-learning strategy to adaptively distill knowledge from all samples during the self-training process.

Alternatively, pseudo-labels can also be used  to improve the model without necessarily using them in a self-training cross-entropy loss. For example, Wang~\etal~\cite{WangCVPR20DifferentialTreatmentStuffThingsDASemSegm}  use them to disentangle 
source and target features by taking into account regions associated with~\textit{things} and \textit{stuff}~\cite{HolgerCVPR2018COCOStuffThingsAndStuffClassesContext}.



Du~\etal~\cite{DuICCV19SSFDANSeparatedSemanticFeatureDASemSegm}
use separate semantic features according to the downsampled pseudo-labels to build class-wise confidence map used to reweigh the adversarial loss. A progressive confidence strategy is used to obtain reliable pseudo-labels and, hence, class-wise confidence maps. 

\myparagraph{Entropy minimization of target predictions (TEM)}
Originally devised for semi-supervised learning~\cite{GrandvaletNIPS2004SSLEntropyMinimization},  entropy minimization has received a broad recognition as an alternative or complementary technique for domain alignment. Different DASiS/UDA methods extend simple entropy minimization on the target data by applying it jointly with adversarial losses~\cite{DuICCV19SSFDANSeparatedSemanticFeatureDASemSegm,VuCVPR19ADVENTAdversarialEntropyMinimizationDASemSegm} or square losses~\cite{ChenICCV19DASemSegmentationMaximumSquaresLoss,ToldoWACV21UDASemSegmOrthogonalClusteredEmbeddings}. 

Vu~\etal~\cite{VuCVPR19ADVENTAdversarialEntropyMinimizationDASemSegm}
propose to enforce structural consistency across domains by minimizing both the conditional entropy of pixel-wise predictions and an adversarial loss that performs the distribution matching in terms of weighted entropy maps. The main advantage of their approach is that computation of the pixel-wise entropy does not depend on any network and entails no overhead. 


Similarly, Huang~\etal~\cite{HuangECCV20ContextualRelationConsistentDASemSegm} design an entropy-based max-min adversarial learning scheme to align local contextual relations across domains.
The model learns to enforce the prototypical local contextual relations explicitly in the feature space of a labelled source domain, while transferring them to an unlabelled target domain 
via backpropagation-based adversarial learning
using a GRL~\cite{GaninJMLR16DomainAdversarialNN}. 


Chen~\etal~\cite{ChenICCV19DASemSegmentationMaximumSquaresLoss} show that entropy minimization based UDA methods often suffer from the probability imbalance problem. To avoid the adaptation process to be dominated by the easiest to adapt samples, they propose instead a class-balanced weighted maximum squares loss with a linear growth gradient.
Furthermore, they extend the objective function of target samples with self-training with low-level features guided by pseudo-labels obtained by averaging the output map at different levels of the network. Toldo~\etal~\cite{ToldoWACV21UDASemSegmOrthogonalClusteredEmbeddings}  integrate this image-wise class-balanced entropy-minimization loss to regularize their feature clustering based DASiS method. To further  enhance the discriminative clustering performance they introduce 
an orthogonality loss forcing individual representations far away to be orthogonal and a sparsity loss to reduce class-wise the number of active feature channels.



\begin{figure}[ttt]
\begin{center}
\includegraphics[width=0.47\textwidth]{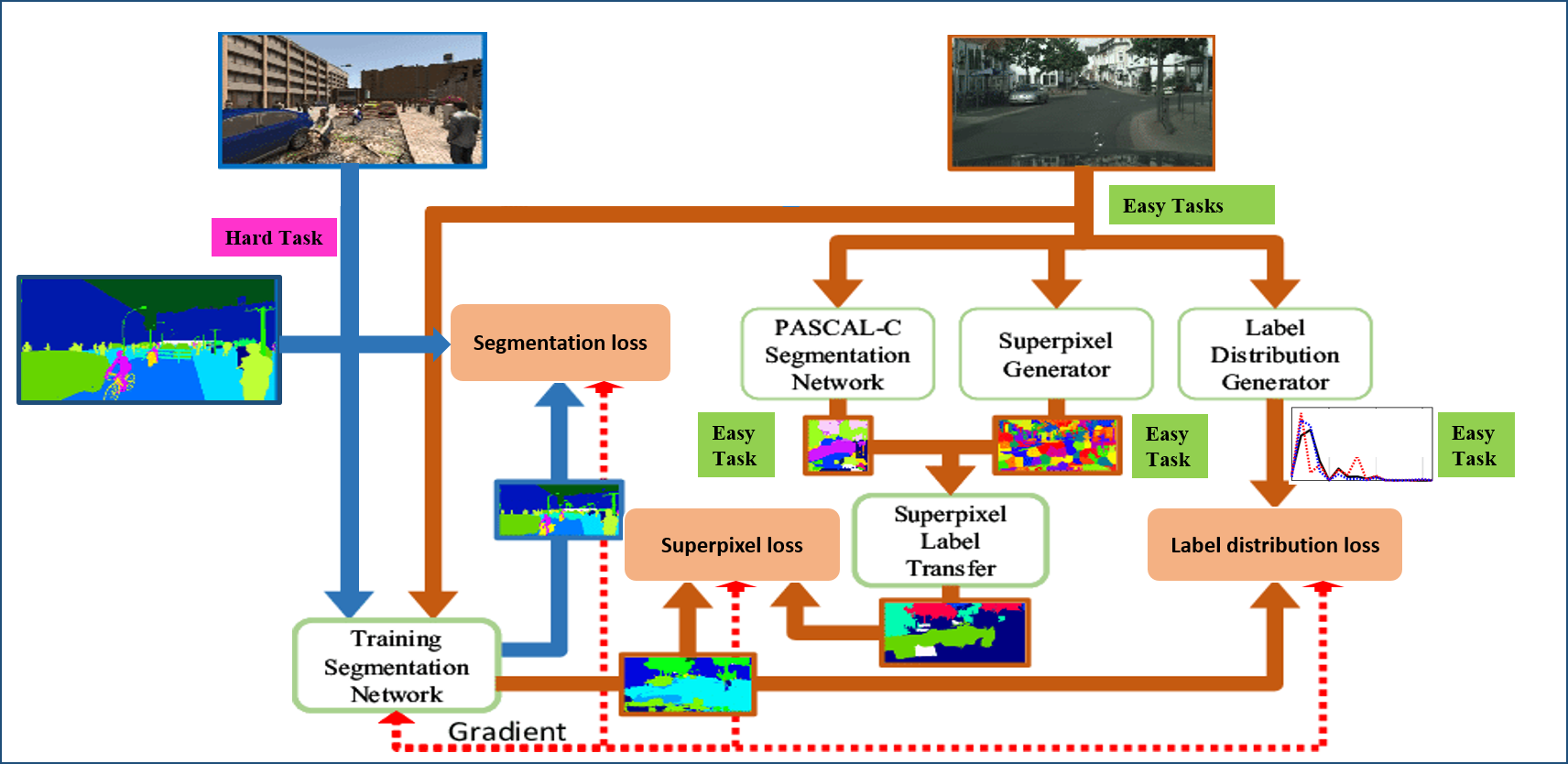}
\caption{Zhang~\etal~\cite{ZhangPAMI20CurriculumDASemSegmUrbanScenes}  propose a curriculum learning based approach to DASiS, where they start from two easy tasks (predicting image-level and superpixel-level label distributions), less sensitive to the domain discrepancy and use them to guide the pixel-level target segmentation.}
\label{fig:CurrDASemSeg}
\end{center}
\end{figure}

Truong~\etal~\cite{TruongICCV21BiMaLBijectiveMaximumLikelihoodDASS} propose a new Bijective Maximum Likelihood (BiMaL) loss that is a generalized form of the adversarial entropy minimization, without any assumption about pixel independence. The BiMaL loss is formed using a maximum-likelihood formulation to model the global structure of a segmentation input and a bijective function to map that segmentation structure to a deep latent space. Additionally, an Unaligned Domain Score (UDS) is introduced to measure the efficiency of the learned model on a target domain in an unsupervised fashion.

\myparagraph{Curriculum learning (CurrL)} 
Curriculum learning refers to the practice of designing the training process such that \textit{easier tasks} are approached first, and \textit{harder tasks} then~\cite{BengioICML2009CurriculumLearning}. Such an approach has been extended to tackle the DASiS formulation by 
Zhang~\etal~\cite{ZhangPAMI20CurriculumDASemSegmUrbanScenes}: they show that the segmentation networks trained on the source domain perform poorly on many target images due to disproportionate label assignments 
To rectify this issue, they propose to use image-level label distribution to guide the  pixel-level target segmentation. Furthermore, they also use the label distributions of  anchor superpixels to indicate the network where to update. After these ``easier tasks'', the predicted pseudo-labels for the target samples are used to effectively regularize the fine-tuning of the SiS network (see~\fig{CurrDASemSeg}).

Similarly, Sakaridis~\etal~\cite{SakaridisICCV19GuidedCurriculumModelAdaptationSemSegm} propose a curriculum learning approach where models are adapted from day to night at increasing level of darkness. They exploit the correspondences across time of different images --- in order to improve pixel predictions at inference time.
Lian~\etal~\cite{LianICCV19ConstructingSelfMotivatedPyramidCurriculumsCrossDomainSemSegm} adopt the easy-to-hard curriculum learning approach by predicting labels first at image level, then at region level and finally at pixel level (the main task). 
Pan~\etal~\cite{PanCVPR20UnsupervisedIntraDASemSegmSelfSupervision} rely on  the entropy to rank target images and to separate them into an easy and a hard split, in order to learn the adaptation 
from clean to noisy predictions within the target domain. 

\myparagraph{Co-training (CoT)}
Another set of UDA/DASiS methods have been inspired by co-training~\cite{ZhouTKDE05TriTrainingExploitingUnlabeledData} where the idea is to have two distinct classifiers enforced to be diverse, in order to capture different views of the data while predicting the same labels. The main idea behind such methods is that diversifying the classifiers in terms of learned parameters --- while at the same time maximizing the consensus on their predictions --- will encourage the model to output more discriminative feature maps for the target domain. The rationale is that the target samples near the class boundaries are likely to be misclassified by the source classifier and using the disagreement of two classifiers on the prediction for target samples can implicitly detect them and, hence, correct the class boundaries. 

The first such UDA model maximizing the classifier discrepancy (MCD) was proposed by Saito~\etal~\cite{SaitoCVPR18MaximumClassifierDiscrepancyUDA} where the adversarial model alternates between (i) maximizing the discrepancy between two classifiers on the target sample while keeping the feature generator fixed and (ii) training the feature encoder to minimize discrepancy while keeping the classifiers fixed.
Similarly, to encourage the encoder to output more discriminative features for the target domain, Saito~\etal~\cite{SaitoICLR18AdversarialDropoutRegularization} rely on adversarial dropout, and  Luo~\etal~\cite{LuoCVPR19TakingACloserLookDomainShiftSemanticsConsistentDA} enforce the weights of the two classifiers to be diverse while using self-adaptive weights in the adversarial loss to improve local semantic consistency. Finally, Lee~\etal~\cite{LeeCVPR19SlicedWassersteinDiscrepancyUDA} consider the sliced Wasserstein discrepancy to capture the dissimilarity between the predicted probability measures which provides a geometrically meaningful guidance to detect target samples that are far from the support of the source.

\begin{figure}[ttt]
\begin{center}
\includegraphics[width=0.47\textwidth]{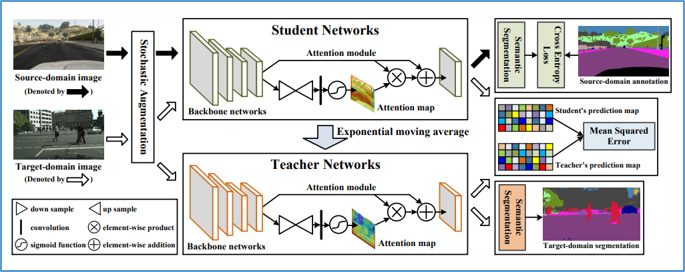}
\caption{ The self-ensembling attention network ~\cite{XuAAAI19SelfEnsemblingAttentionNetworksSemSegm} consists of a student network and a teacher network. The two networks share the same architecture with an embedded attention module. The student network is jointly optimized with a supervised segmentation loss for the source domain and an unsupervised consistency loss for the target domain. The teacher network is excluded from the back-propagation, it is updated with an exponential moving average. In the test phase, the target-domain images are sent to the teacher network to accomplish the SiS task.}
\label{fig:self_ens}
\end{center}
\end{figure}

\myparagraph{Self-ensembling}
Another popular method for semi-supervised learning is to use an ensemble of models and exploit the consistency between predictions under some perturbations. 
Laine~\etal~\cite{LaineX16TemporalEnsemblingSemisupervisedLearning} propose the temporal ensembling of per-sample moving average of predictions; Tarvainen~\etal~\cite{TarvainenNIPS17MeanTeachersBetterRoleModelsSSL} replace the averaging predictions with an exponential moving average (EMA) of the model weights: They propose a Mean Teacher framework with a second, non-trainable model, whose weights are updated with the EMA over the actual trainable weights.

Such self-ensembling models can also be used for UDA and DASiS, where the model is generally composed of a teacher and a student network --- encouraged to produce consistent predictions. The teacher is often an ensembled model that averages the student’s weights and, hence, the predictions from the teacher can be interpreted as pseudo-labels for the student model.
Indeed, French~\etal~\cite{FrenchICLR18SelfEnsemblingForVisualDA} extend the model proposed by Tarvainen~\etal~\cite{TarvainenNIPS17MeanTeachersBetterRoleModelsSSL} to UDA considering a separate path for source and target, and sampling independent batches making the Batch Normalization (BN)~\cite{IoffeICML15BatchNormAcceleratingDeepNetsTrain} domain specific during the training process. Perone~\etal~\cite{PeroneNI19UDAMedicalImSegmSelfEnsembling} apply self-ensembling to adapt medical imaging segmentation. The Self-ensembling Attention Network of Xu~\etal~\cite{XuAAAI19SelfEnsemblingAttentionNetworksSemSegm} aims at extracting attention aware features for domain adaptation.

In contrast to the above mentioned ensemble models, which are efficient but require heavily-tuned manual data augmentation for successful domain alignment,  Choi~\etal~\cite{ChoiICCV19SelfEnsemblingGANBasedDataAugmentationDASemSegm} propose a self-ensembling framework which deploys a target-guided GAN-based data augmentation with spectral normalization. 
To produce semantically accurate prediction for the source and augmented samples, a semantic consistency loss is used.
More recently, Wang~\etal~\cite{WangAAAI21ConsistencyRegularizationSourceGuidedPerturbationUDASemSegm} propose a method that relies on AdaIN~\cite{HuangICCV17ArbitraryStyleTransferRealTimeAdaptiveInstanceNormalization} to convert the style of source images into that of the target domain, and vice-versa. The stylized images are exploited in training pipeline that exploits self-training, where pseudo-labels are improved via the usage of self-ensembling.

\begin{figure}[ttt]
\begin{center}
\includegraphics[width=0.47\textwidth]{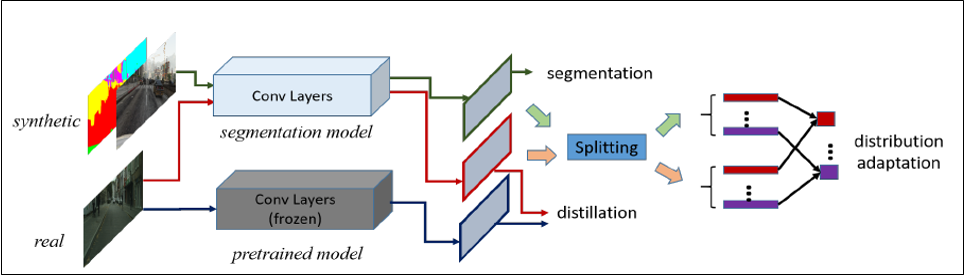}
\caption{Chen~\etal~\cite{ChenCVPR18RoadRealityOrientedDASemSegmUrbanScenes} propose to incorporate a target guided distillation module to learn the target (real) style convolutional filters from the (synthetic) source ones and combine it with a spatial-aware distribution adaptation module.}
\label{fig:ROAD}
\end{center}
\end{figure}

\myparagraph{Model distillation}
In machine learning, ``\textit{model distillation}''~\cite{HintonX15DistillingKnowledgeNN} has
been introduced as a way to transfer the knowledge from a large model to a smaller one --- for example, compressing the discriminative power of an ensemble of models into a single, lighter one. In the context of DASiS, it has been exploited to guide the learning of more powerful features for the target domain, transferring the discriminative power gained on the source samples.

Chen~\etal~\cite{ChenCVPR18RoadRealityOrientedDASemSegmUrbanScenes} propose to tackle the distribution alignment
in DASiS by using a distillation strategy to learn the target style convolutional filters (see \fig{ROAD}).
Furthermore, taking advantage of the intrinsic spatial structure presented in urban scene images (that they focus on), they propose to partition the images into non-overlapping grids, and the domain alignment is performed on the pixel-level features from the same spatial region using GRL~\cite{GaninJMLR16DomainAdversarialNN}.
Finally, The Domain Adaptive Knowledge Distillation model~\cite{KothandaramanWACV21DomainAdaptiveKnowledgeDistillationDrivingSceneSemSegm} consists in a multi-level strategy to effectively distill knowledge at different levels --- feature space and output space --- using a combination of KL divergence and MSE losses.

\myparagraph{Adversarial attacks}
The aim of adversarial attacks \cite{SzegedyICLR14IntriguingPropertiesNeuralNetworks} is to
perturb examples in a way that makes deep neural networks fail when processing them. Training with both clean and adversarial samples is a technique called ``\textit{adversarial training}'' \cite{GoodfellowICLR15ExplainingHarnessingAdvExamples}, which effectively allows to learn more robust models for the given task. 
While the connection between adversarial robustness and generalization is not fully explained
yet~\cite{GilmerICML19AdvExamplesNaturalConsequenceTestErrNoise}, adversarial training has been successfully applied to achieve different goals than adversarial robustness; for instance, it has been used to mitigate over-fitting in supervised and semi-supervised
learning~\cite{ZhengCVPR16ImprovingRobustnessDeepNNStabilityTraining}, to tackle domain generalization tasks~\cite{VolpiNeurIPS19GeneralizingUnseenDomainsAdvDataAugm}, or to fill in the gap between the source and target domains and adapt the classification decision boundary
\cite{LiuICML19TransferableAdversarialTrainingGeneralApproach}.

Concerning DASiS, Yang~\etal~ \cite{YangAAAI20AnAdversarialPerturbationOrientedDASS} propose pointwise perturbations to  generate adversarial features that capture the vulnerability of the model (\eg, the tendency of the classifier to collapse into the classes that are more represented, in contrast with the long tail of the most under-represented ones) and conduct adversarial training on the segmentation network to improve its robustness. 
Yang~\etal~\cite{YangICCV21ExploringRobustnessDASS} study the adversarial vulnerability of existing DASiS methods and propose the adversarial self-supervision UDA, where the objective is to maximize the proximity between clean images and their adversarial counterparts in the output
space --- by using a contrastive loss. 
Huang~\etal~\cite{HuangICCV21RDARobustDAViaFourierAdversarialAttacking} propose a Fourier adversarial training method, where the pipeline is (i) generating adversarial samples by perturbing certain high frequency components that do not carry significant semantic information and (ii) using them to train the model. This training technique allows reaching an area with a flat loss landscape, which yields a more robust domain adaptation model.

\begin{figure*}[ttt]
\begin{center}
\fbox{\includegraphics[width=0.85\textwidth]{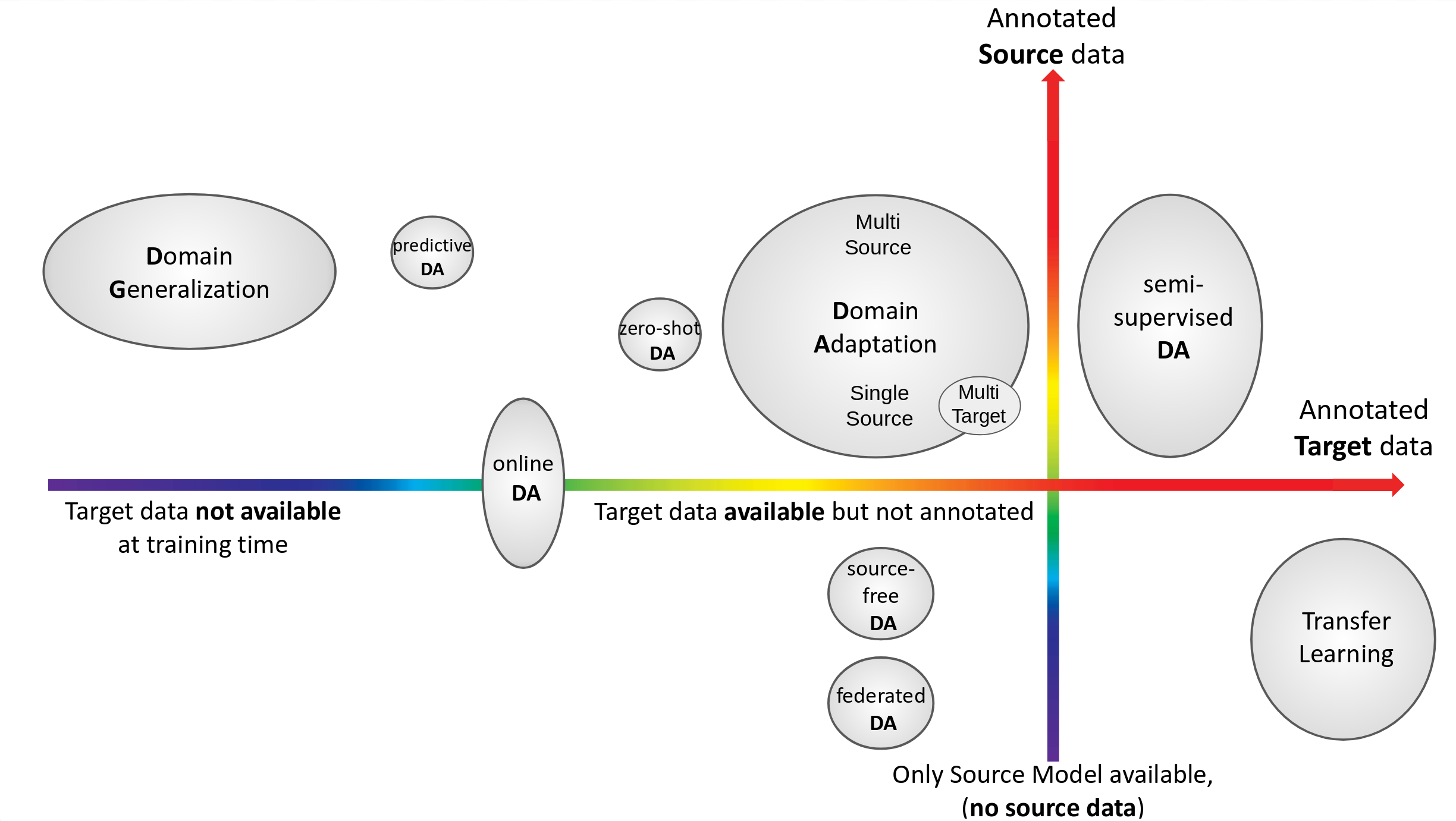}}
\caption{Illustration of different scenarios based on source and target data availability. (Image courtesy of Tatiana Tommasi). }
\label{fig:beyondDA}
\end{center}
\end{figure*}

\myparagraph{Self-supervised learning}
In the absence of human annotations, self-supervised learning approaches represent a very effective alternative to learn strong visual representations. The idea is devising an auxiliary task, such as rotation prediction~\cite{GidarisICLR18UnsupervisedRepresentationLearningPredictImageRotations},
colorization~\cite{ZhangECCV16ColorfulImageColorization}, or contrastive learning~\cite{ChenICML20SimpleFrameworkContrastiveLearningVisualRepresentations}, and train the model to solve this task instead of a supervised one. Self-supervised learning approaches have found their place in UDA research,~\eg see~\cite{SunX19UDAThroughSSL}.

For what concerns DASiS,
Araslanov~\etal~\cite{AraslanovCVPR21SelfSupervisedAugmentationConsistencyAdaptingSemSegm} propose a lightweight self-supervised training scheme, where the consistency of the semantic predictions across image transformations such as photometric noise, mirroring and scaling is ensured.
The model is trained end-to-end using co-evolving pseudo-labels --- using a momentum network (a copy of the original model that evolves slowly) --- and maintaining an exponentially moving class prior, which is used to discount the confidence thresholds for classes with few samples, in order to  increase their relative contribution to the training loss.
Also, Yang~\etal~\cite{YangICCV21ExploringRobustnessDASS} --- as mentioned in the previous paragraph --- exploit self-supervision in DASiS by minimizing the distance between clean and adversarial samples in the output space
via a contrastive loss.

\section{Beyond classical DASiS}\label{sec:beyond_dass}

Typical 
DASiS methods assume that both source and target domains
consist of samples drawn from single data distributions, both available, and that there is a shift between the two distributions. Yet, these assumptions may not hold in the real world, and hence several methods have been proposed that tackle specific problem formulations where some of these assumptions are relaxed (see~\fig{beyondDA} for an illustration of different scenarios related to different data availability assumptions). 

For instance, in {\it multi-source} domain adaptation (MSDA) the goal is learning from an arbitrary number of source domains (Section~\ref{ssec:msdass}), and in {\it multi-target} domain adaptation (MTDA) the aim is to learn from a single source and several unlabeled target domains (Section~\ref{ssec:mtdass}). Another scenario is {\it domain generalization}, where the model learns from one or multiple source domains, but has not access to any target sample, nor hints on the target distribution (Section~\ref{ssec:dg}). On the other end, different methods tackle the semi-supervised domain adaptation problem, where one even assumes that a few target samples are annotated (Section~\ref{s-SSDA}), or can be actively annotated (Section~\ref{sec:act-DASS}). One could finally make the assumption that the source model is available, but the source data it was trained on is not --- the source-free domain adaptation problem (Section~\ref{s-SFDA}).

Besides the number of domains and the amount of labeled/unlabeled samples available in the source/target domains, another important axis of variation for domain adaptation strategies is the label overlap between source and target. Indeed, the class of semantic labels in the source and the target domains is not necessarily the same, and hence several methods have been proposed that address this issue  
(Section~\ref{ssec:universal}).

\subsection{Multi-source DASiS} 
\label{ssec:msdass}

The simplest way to exploit multiple source domains is to combine all the available data in a single source and train a classical UDA model. While this can in some cases  provide a reasonable baseline, in other cases it might yield poor results. This can be due to 
1) the fact that there are several data distributions mixed in the combined source, making the adaptation process more difficult if this is not explicitly handled and 2) in many cases this might yield to strong  negative transfer as shown  in \cite{MansourNIPS09DAMultipleSources}.
Alternatively, one can consider  a weighted combination of multiple source domains for which theoretical analysis of error bounds was proposed in~\cite{BenDavidML10ATheoryLearningDifferentDomains,CrammerJMLR08LearningFromMultipleSources}. 
Hoffman~\etal~\cite{HoffmanNIPS18AlgorithmsTheoryMultipleSourceAdaptation} propose an algorithm with strong theoretical guarantees to determine the distribution-weighted combination solution for the cross-entropy loss and other similar losses. Cortes~\etal~\cite{CortesICML21ADiscriminativeTechniqueMSDA} propose instead a discriminative method which only needs conditional probabilities  that can be accurately estimated  for the unlabeled target data, relying only on the access to the source predictors and not the labeled source data. 
Russo~\etal~\cite{RussoICIAP19TowardsMultiSourceAdaptiveSemSegm} extend adversarial DASiS
to deal with multiple sources and investigate such baselines, \ie comparing models trained on the union of the source domains versus weighted combination of adaptive adversarial models trained on individual source-target pairs.

Further methods proposed for image classification \cite{LiNIPS18ExtractingRelationshipByMultiDomainMatching,PengICCV19MomentMatchingMultiSourceDA,PengECCV20Domain2VecDomainEmbeddingUDA,YangECCV20CurriculumManagerForSourceSelectionMSDA,ZhaoNIPS18AdversarialMultipleSourceDA,ZhaoAAAI20MultiSourceDistillingDA,ZhouX20DomainAdaptiveEnsembleLearning,ZhuAAAI19AligningDomainSpecificDistributionMSDA,NguyenICCV21STEMMultiSourceDAWithGuarantees} show that when the relationship between different source domains is appropriately exploited, it is possible to train a target model which performs significantly better than using just the union of source data or a weighted combination of individual models' outputs. These deep multi-source DA (MSDA) approaches have often focused on learning a common domain-invariant feature extractor that achieves a small error on several source domains hoping that such representation  can generalize well to the target domain.

\begin{figure}[ttt]
\begin{center}
\includegraphics[width=0.47\textwidth]{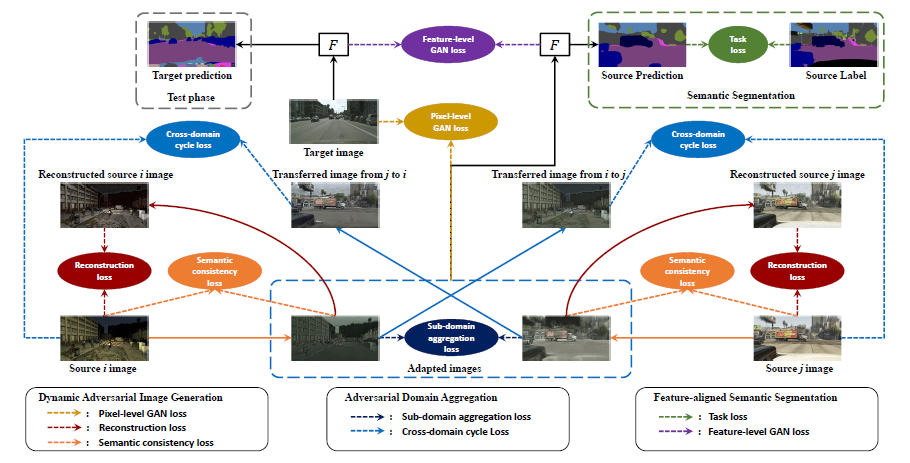}
\caption{The multi-source DASiS framework proposed by Zhao~\etal~\cite{ZhaoNIPS19MultiSourceDASemSegm}  
consists in adversarial domain aggregation 
with two kinds of discriminators: a sub-domain aggregation discriminator, which is
designed to directly make the different adapted domains indistinguishable, and a cross-domain cycle discriminator, discriminates between the images from each source and the images transferred from other sources.}
\label{fig:MSDASiS}
\end{center}
\end{figure}

Inspired by these approaches, several methods were proposed  
that extend MSDA solutions from classification to semantic image segmentation   \cite{ZhaoNIPS19MultiSourceDASemSegm,TasarCWPRWS20MSDASemSegmSatelliteImagesDataStandardization}.
As such, the Multi-source Adversarial Domain Aggregation Network~\cite{ZhaoNIPS19MultiSourceDASemSegm} 
extends \cite{ZhaoNIPS18AdversarialMultipleSourceDA} 
by combining it with CyCADA~\cite{HoffmanICML18CyCADACycleConsistentAdversarialDA}. The model, trained end-to-end, generates for each source an adapted style transferred domain
with dynamic semantic consistency loss between the source predictions of a pre-trained segmentation model and the adapted predictions of a dynamic segmentation model. To make these adapted domains indistinguishable, a sub-domain aggregation discriminator and a cross-domain cycle discriminator is learned in an adversarial manner (see~\fig{MSDASiS}).

Similarly, Tasar~\etal~\cite{TasarCWPRWS20MSDASemSegmSatelliteImagesDataStandardization} propose  StandardGAN, a data standardization technique based on GANs (style transfer) for satellite image segmentation, whose goal is to standardize the visual appearance of the source and the target domain with adaptive instance normalization  \cite{HuangICCV17ArbitraryStyleTransferRealTimeAdaptiveInstanceNormalization} and LSGAN~\cite{MaoICCV17LeastSquaresGAN}
to effectively process target samples. Then, they extend the single-source StandardGAN to multi-source 
by multi-task learning where an auxiliary classifier is added on top of the discriminator.

In contrast, 
He~\etal~\cite{He2021CVPRMultiSourceDACollaborativeLearningforSS} propose a collaborative learning approach. They first translate source domain images to the target style by aligning the different distributions to the target domain in LAB color space. Then, the SiS network for each source is trained in a supervised fashion by relying on GT annotations and additional soft supervision coming from other
models trained on a different source domain. Finally, the segmentation models associated with different sources collaborate with each other to generate more reliable pseudo-labels for the target domain --- finally used to refine the models.

Gong~\etal~\cite{GongICCV21mDALUMuSDALabelUnificationWithPartialDatasets}  consider the case where the aim is to learn from different source datasets with potentially different class sets, and formulates it as a multi-source domain adaptation  
with ``\textit{label unification}''. To approach this, they propose a two-step solution: first, the knowledge is transferred form the multiple sources to the target;  second, a unified label space is created by exploiting pseudo-labels, and the knowledge is further transferred to this representation space.  To address in the first step the risk of making confident predictions for unlabeled samples in the source domains and, hence, to enable robust distribution alignment between the source domains and the target domain, three novel modules are proposed ---  domain attention, uncertainty maximization and attention-guided adversarial alignment.

\subsection{Multi-target DASiS} 
\label{ssec:mtdass}

In multi-target domain adaptation (MTDA) the goal is learning from a single labeled source domain 
with the aim of performing well on multiple target domains --- at the same time. To tackle MTDA 
within an image classification context, standard UDA approaches can be directly
extended to multiple targets~
\cite{GholamiTIP20UnsupervisedMultiTargetDAInformationTheoreticApproach,ChenCVPR19BlendingTargetDAAdversarialMetaAdaptationNetworks,RoyCVPR21CurriculumGraphCoTeachingMTDA,NguyenMeidineWACV21UnsupervisedMultiTargetDAKnowledgeDistillation}. 


Within the DASiS context, different approaches have been proposed~\cite{IsobeCVPR21MTDAWithCollaborativeConsistencyLearning,SaportaICCV21MultiTargetAdversarialFrameworksForDASS}. 
Isobe~\etal~\cite{IsobeCVPR21MTDAWithCollaborativeConsistencyLearning} propose a method that trains an expert model for every target domain where the models are encouraged to collaborate via style transfer. 
Such expert models are further exploited as teachers for a student model that learns to imitate their output and serves as regularizer to bring the different experts closer to each other in the learned feature space. Instead,
Saporta~\etal~\cite{SaportaICCV21MultiTargetAdversarialFrameworksForDASS} propose to combine for each target domain $\calT_i$ two adversarial pipelines: one that learn to discriminate between the domain $\calT_i$ and the source,  and one 
between $\calT_i$ and the union of other target domains. 
 Then, to reduce the instability that the multi-discriminator model training might cause, they propose a  multi-target knowledge transfer by adopting a multi-teacher/single-student distillation mechanism, which leads to a model that is agnostic to the target domains.

\begin{figure}[ttt]
\begin{center}
\includegraphics[width=0.4\textwidth]{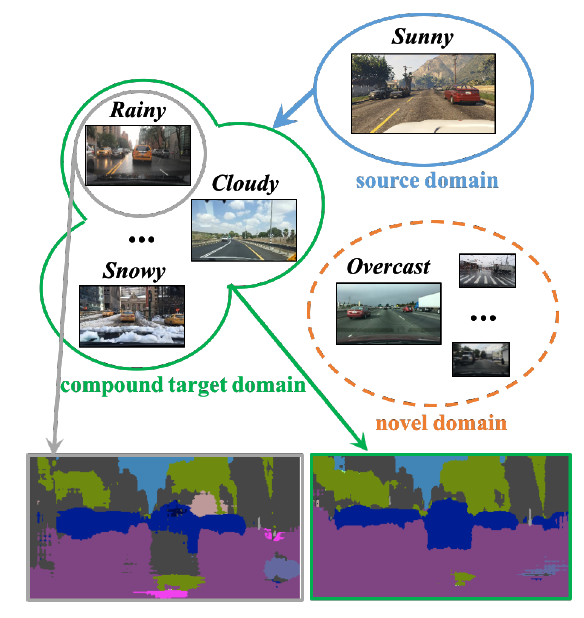}
\caption{In the Open Compound Domain Adaptation (OCDA) setting, the target distribution 
is assumed to be a compound of multiple, unknown, homogeneous domains (Illustration from Liu~\etal~\cite{LiuCVPR20OpenCompoundDA}).}
\label{fig:OpenCompoundDA}
\end{center}
\end{figure}

The possibility of having multiple target domains is also addressed in
the \textit{Open Compound Domain Adaptation} (OCDA) setting, where the target distribution 
is assumed to be a compound of multiple, unknown, homogeneous domains (see \fig{OpenCompoundDA}).
To face this problem, Liu~\etal~\cite{LiuCVPR20OpenCompoundDA} rely on a curriculum adaptive strategy, where they schedule the learning of unlabeled instances in the compound target domain according to their
individual gaps to the labeled source domain, approaching an incrementally harder and harder domain adaptation problem until the entire target domain is covered.
The purpose of this is to learn a network that maintains its discriminative leverage on the classification or segmentation  task at hand, while at the same time learning more robust features for the whole compound domain. To further prepare the model for open domains during inference, a memory module is adopted to effectively augment the representations of an input when it is far away from the source. 
In contrast, to approach OCDA, Gong~\etal~\cite{Gong2021CVPRClusterSplitFuseUpdate} 
propose a meta-learning based approach 
where, first, the target domain is clustered in an unsupervised manner into multiple sub-domains by image styles; 
then, different sub-target domains are split into independent branches, for which domain-specific BN parameters are learned 
as in \cite{ChangCVPR19DomainSpecificBNforUDA}. Finally, a meta-learner is deployed to learn to fuse sub-target domain-specific predictions, conditioned upon the style code, which  
is updated online by using the model-agnostic meta-learning 
algorithm that further improves its generalization ability.

\subsection{Domain generalization}
\label{ssec:dg}
Domain generalization (DG) aims to generalize a model trained on one or several source domains to new, unseen target domains and therefore its main goal is to learn {\it domain-agnostic representations}. As the target domain is unknown at training time, most DG methods aim to minimize the average risk over ``\textit{all possible}'' target domains. According to \cite{WangX20GeneralizingToUnseenDomainsSurveyDG,ZhouX20DomainGeneralizationSurvey}, such DG methods can be categorized into multiple groups, relying on data randomization~\cite{TobinIROS17DomainRandomizationTransferringDNNSimToReal}, 
ensemble learning~\cite{XuECCV14ExploitingLowRankDomainGeneralization,SeoECCV20LearningToOptimizeDomainSpecificNormalizationDG}, meta-learning~\cite{BalajiNIPS19MetaRegTowardsDGMetaRegularization,LiICML19FeatureCriticNetworksHeterogeneousDG,RahmanPR20CorrelationAwareAdversarialDADG} domain-invariant representation learning~\cite{LiCVPR18DomainGeneralizationWithAdversarialFeatureLearning,MotiianICCV17UnifiedDeepSupervisedDADG}, feature disentanglement~\cite{ChattopadhyayECCV20LearningToBalanceSpecificityInvarianceDG}, self-supervised learning\cite{CarlucciCVPR19DomainGeneralizationBySolvingJigsawPuzzles,WangECCV20LearningFromExtrinsicIntrinsicSupervisionsDG}, invariant risk minimization~\cite{ArjovskyX20InvariantRiskMinimization} and others.

\begin{figure}[ttt]
\begin{center}
\includegraphics[width=0.48\textwidth]{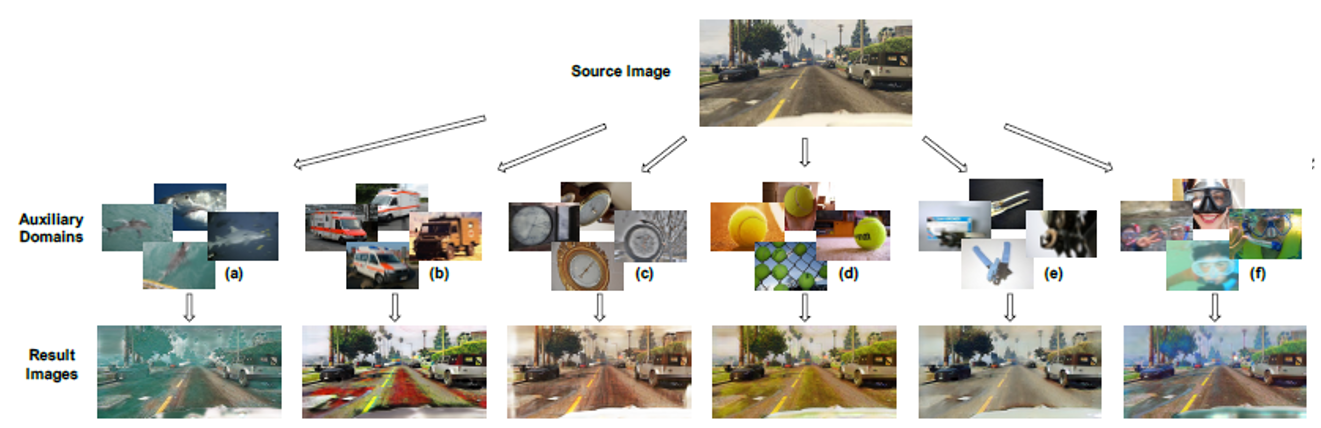}
\caption{The domain randomization process proposed by Yue~\etal~\cite{YueICCV19DomainRandomizationPyramidConsistencySimulationToReal}, where auxiliary image sets corresponding to ImageNet classes are used to stylize the target images.}
\label{fig:DomRand}
\end{center}
\end{figure}

Concerning DG methods proposed for SiS (DGSiS), different techniques were similarly explored. Yue~\etal~\cite{YueICCV19DomainRandomizationPyramidConsistencySimulationToReal} rely on domain randomization, where --- using auxiliary datasets --- the synthetic images are translated with multiple real image styles to effectively learn domain-invariant and scale-invariant representations (see \fig{DomRand}).

Gong~\etal~~\cite{GongCVPR19DLOWDomainFlowAdaptationGeneralization} propose to learn domain-invariant representations via domain flow generation. The main idea is to generate  a continuous sequence of intermediate domains between the source and the target in order to bridge the gap between them. To translate images from the source domain into an arbitrary intermediate domain, an adversarial loss is used to control how the intermediate domain is related to the two original ones (source and target). Several intermediate domains of this kind are generated, so that the discrepancy between the two domains is gradually reduced in a manifold space. Jin~\etal~\cite{JinX20StyleNormalizationRestitutionDGDA} rely on  style normalization and restitution module to enhance the generalization capabilities, while preserving the discriminative power of the networks for effective DG. The style normalization 
is performed by instance normalization to filter out the style variations and, hence, foster generalization. To ensure high discriminative leverage, a restitution step adaptively distills task-relevant discriminative features from the residual (\ie the difference between original and style normalized features) 
which are then exploited to learn the network. 
Liu~\etal~\cite{LiuTMI20MSNetMultiSiteNetworkProstateSegmHeterogeneousMRIData} extend domain-specific BN layers proposed in \cite{SeoECCV20LearningToOptimizeDomainSpecificNormalizationDG} for MRI image segmentation, where at inference time an ensemble of prediction is generated and  their confidence-weighted average is considered as the final prediction. 
 
While not devised ad-hoc for SiS,~\cite{QiaoCVPR20LearningToLearnSingleDG,VolpiICCV19AddressingModelVulnerabilityDistrShiftsImageTransf,VolpiNeurIPS19GeneralizingUnseenDomainsAdvDataAugm} show that worst-case data augmentation strategies can improve robustness of segmentation models. Indeed, Volpi~\etal~\cite{VolpiNeurIPS19GeneralizingUnseenDomainsAdvDataAugm} propose to create fictitious visual domains that are “hard” for the model at hand, by leveraging adversarial training to augment the source domain, and use them to
train the segmentation model. Qiao~\etal~\cite{QiaoCVPR20LearningToLearnSingleDG} extend this idea by relying on meta-learning; to encourage out-of-domain augmentations, the authors rely on a Wasserstein auto-encoder which is jointly learned with the segmentation and domain augmentation within a meta-learning framework.
Volpi and Murino~\cite{VolpiICCV19AddressingModelVulnerabilityDistrShiftsImageTransf} instead rely on standard image transformations, using random and evolution search to find worst-case perturbations that are further used as data augmentation rules.

Closely related with DGSiS, 
Lengyel~\etal~\cite{LengyelICCV21ZeroShotDayNightDAWithPhysicsPrior} propose ``\textit{zero-shot day-to-night domain adaptation}''  to improve performance on unseen illumination conditions without the need of accessing target samples. The proposed method relies on task agnostic physics-based illumination priors where a trainable Color Invariant Convolution  layer is used to transform the input to a domain-invariant representation. It is shown that this layer allows reducing the day-night domain shift in the feature map activations throughout the network and, in turn, improves SiS on samples recorded at night.

\subsection{Semi-supervised domain adaptation}
\label{s-SSDA}

Semi-supervised learning (SSL) methods exploit at training time accessibility to both a small amount of labeled data and a large amount of unlabeled data. After gaining traction for more standard classification tasks, recently several semi-supervised methods have emerged that address SiS problems.
Some methods rely on GAN-inspired adversarial losses that can be trained on the unlabeled data~\cite{SoulyICCV17SemiSupSemSegmGAN,HungBMVC18AdversarialLearning4SemiSupervisedSemSegm,MittalPAMI21SemiSupSemSegmHighLowLevelConsistency}; other methods exploit data augmentation and consistency regularization~\cite{FrenchBMVC19SemiSupervisedSemSegmHighDimPerturbations,OualiCVPR20SemiSupervisedSemSegmCrossConsistencytraining,LuoECCV20SemiSupervisedSemSegmStrongWeakDualBranchNetwork,YuanICCV21ASimpleBaselineSemiSupSemSegmStrongDataAugmentation}.

The standard UDA setting shares with semi-supervised learning the availability at training time of labeled and unlabeled data; the core difference is that in the semi-supervised framework both sets are drawn from the same domain (\iid assumption), whereas they are drawn from different data distributions --- source and target --- in UDA.
In Section~\ref{sec:dass_improvements} we have discussed how several strategies from the SSL literature such as pseudo-labeling, self-training, entropy minimization, self-ensembling, have been inherited by DASiS and tailored for cross-domain tasks.

Semi-supervised domain adaptation can be seen as a particular case of them, where on the one hand we can see part of pseudo-labels replaced by GT target labels,  or on the other hand we can see the source labelled data extended with labelled target samples.

To address such scenario, Wang~\etal~\cite{WangCVPRW20AlleviatingSemLevelShiftSSDAForSemSegm} 
leverage a few labeled images from the target domain to supervise the segmentation task and the adversarial semantic-level feature adaptation. 
They show that the proposed strategy improves also over a target domain's oracle.
Chen~\etal~\cite{Chen2021SemiSupervisedDomainAdaptation} tackle the semi-supervised DASiS problem with a method that relies on a variant of CutMix~\cite{YunICCV19CutMixRegStrategyToTrainStrongClassifWithLocFeats} and a student-teacher approach based of self-training. Two kinds of data mixing methods are proposed: on the one hand, 
directly mixing  labeled images from two domains from holistic view; on the other hand, region-level data mixing is achieved by applying two masks to labeled images from the two domains to encourage
the model to extract domain-invariant features about semantic structure from partial view. Then, a student model is trained by distilling knowledge from the 
two complementary domain-mixed teachers --- one obtained by direct mixing and another obtained by region-level data mixing --- and which is refined in a self-training manner for another few rounds of teachers trained with pseudo-labels.

\begin{figure}[ttt]
\begin{center}
\includegraphics[width=0.47\textwidth]{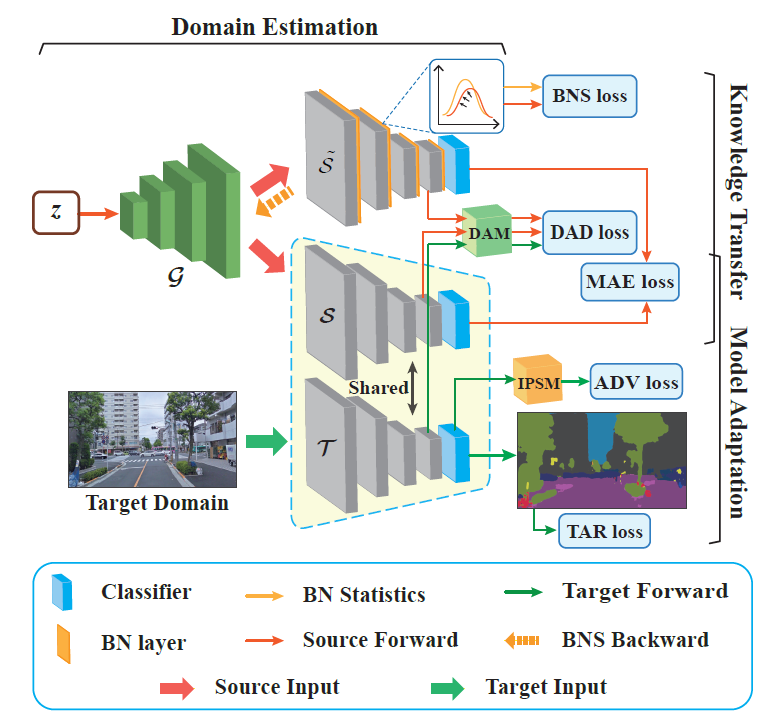}
\caption{Illustration of the source-free DASiS method proposed by  Liu~\etal~\cite{Liu2021SourceFreeDASS} where during training only a well-trained source model and the unlabeled target domain set are accessible --- not the source data. To tackle the problem they propose a dual attention distillation mechanism (DAM) to help the generator synthesize samples with meaningful semantic context, which is beneficial to efficient pixel-level domain knowledge transfer. Furthermore, they rely on   an entropy-based intra-domain patch-level self-supervision module (IPSM) to leverage the correctly segmented patches as self-supervision during the model adaptation stage. }
\label{fig:SFDA}
\end{center}
\end{figure}

\subsection{Active DASiS}\label{sec:act-DASS}

Related to semi-supervised DASiS is the active DASiS
\cite{NingICCV21MultiAnchorActiveDASS,ShinICCV21LabORLabelingOnlyIfRequiredDASS} where, instead of assuming that a small set of target samples are already labeled, an algorithm selects itself the images or pixels to be annotated by human annotators and use them to update the segmentation model over iterations --- in a semi-supervised fashion. Ning~\etal~\cite{NingICCV21MultiAnchorActiveDASS} propose a multi-anchor based active learning strategy to identify the most complementary and representative samples for manual annotation by exploiting the feature distributions across the target and source domains. Shin~\etal~\cite{ShinICCV21LabORLabelingOnlyIfRequiredDASS} 
-- inspired by the maximization classifier discrepancy (MCD) for DA  \cite{SaitoCVPR18MaximumClassifierDiscrepancyUDA} --- propose a method that selects the regions to be annotated  based on the mismatch in predictions across the two classifiers.

\subsection{Source-free domain adaptation} \label{s-SFDA}

Source-free domain adaptation constitutes the problem of adapting a given source model to a target domain, but without access to the original source dataset. It has been introduced by
Chidlovskii~\etal~\cite{ChidlovskiiACMKDD16DAInTheAbsenceOfSourceDomainData}, who propose solutions for both supervised and unsupervised domain adaptation, testing them in a variety of machine learning problems (\eg, document analysis, object classification, product review classification).

More recently, Li~\etal~\cite{LiCVPR20ModelAdaptationUDAWithoutSourceData} propose to exploit the pre-trained source model as a starting component for an adversarial generative model that generates target-style samples, improving the classifier performance on the target domain --- and, in turn, improving the generation process.
Liang~\etal~\cite{LiangICML20DoWeReallyNeedAccessSourceUDA} learn a target-specific feature extraction module by implicitly aligning target representations to the source hypothesis, with a method that exploits at the same time information maximization and self-training. 
Kurmi~\etal~\cite{KurmiWACV21DomainImpressionSourceDataFreeDA}
treat the pre-trained source model as an energy-based function, in order to learn the joint distribution, and  train a GAN that generates annotated samples that are used throughout the adaptation procedure, without the need of accessing source samples. 
Xia~\etal~\cite{XiaICCV21AdaptiveAdversarialNetworkSourceFreeDA} propose first to rely on a learnable target classifier that improves the recognition ability on source-dissimilar target features, and then to perform adversarial domain-level alignment and contrastive matching --- at category level.

For semantic segmentation, Liu~\etal~\cite{Liu2021SourceFreeDASS} propose a dual attention distillation mechanism to help the generator to synthesize samples with meaningful semantic context used to perform efficient pixel-level domain knowledge transfer and  rely on  an entropy-based intra-domain patch-level self-supervision module to leverage the correctly segmented patches as self-supervision during the model adaptation stage (see \fig{SFDA}). 

Sivaprasad~\etal~\cite{SivaprasadCVPR21UncertaintyReductionDASS} propose a solution where the uncertainty of the target domain samples' predictions is minimized, while the robustness against noise perturbations in the feature space is maximized.
Kundu~\etal~\cite{KunduICCV21GeneralizeThenAdaptSourceFreeDASS} decompose the problem into performing first source-only domain generalization and then adapting the model to the target by self-training with reliable target pseudo-labels.

\begin{figure}[ttt]
\begin{center}
\includegraphics[width=0.47\textwidth]{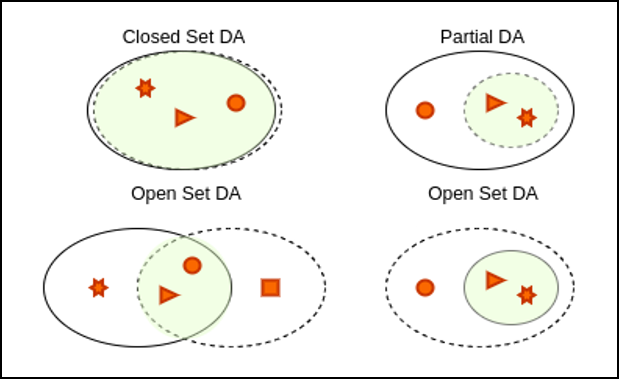}
\caption{A summary of the standard domain adaptation (also known as closed set), partial DA and open set DA  with respect to the overlap between the label sets of the source and the target domains (green shades indicate shared label space). }
\label{fig:OpenSetPartial}
\end{center}
\end{figure}

\subsection{Class-label mismatch across domains} 
\label{ssec:universal}

Another way of sub-dividing domain adaptation approaches is  by considering the mismatch between source and target class sets. Specifically, in \emph{partial domain adaptation}~\cite{ZhangCVPR18ImportanceWeightedAdversarialPartialDA,CaoECCV18PartialAdversarialDA,CaoCVPR19LearningtoTransferExamplesPartialDA} the class set of the source is a super-set of that of the target, while \emph{open set domain adaptation} \cite{PanaredaBustoICCV17OpenSetDomainAdaptation,SaitoECCV18OpenSetDAbyBackpropagation,RakshitECCV20MultiSourceOpenSetDeepAdvDA,JingICCV21TowardsNovelTargetDiscoveryOpenSetDA} assumes that extra private classes exist in the the target domain. Finally, \emph{universal domain adaptation} \cite{FuECCV20LearningToDetectOpenClassesUniversalDA,LiCVPR21DomainConsensusClusteringUniversalDA,SaitoICCV21OVANetOneVsallNetworkUniversalDA,MaICCV21ActiveUniversalDA} integrates both open set and partial DA. 

For what concerns segmentation, Gong~\etal~\cite{GongICCV21mDALUMuSDALabelUnificationWithPartialDatasets} propose a multi-source domain adaptation strategy where the label space of the target domain is defined as the union of the label
spaces of all the different source domains and the knowledge in different label spaces is transferred from different source domains to the target domain and where the missing  labels  are replaced by  pseudo-labels.

Liu~\etal~\cite{LiuICCV21AdvUDAInferAlignIterate}
propose an optimization scheme which alternates between conditional distributions alignment with adversarial UDA relying on estimated class-wise balancing in the target, and target label proportion estimates with mean matching \cite{GrettonBC09CovariateShiftKernelMeanMatching}
assuming conditional distributions alignment between the domains. 


\begin{figure*}[ttt]
\begin{center}
\includegraphics[width=0.9\textwidth]{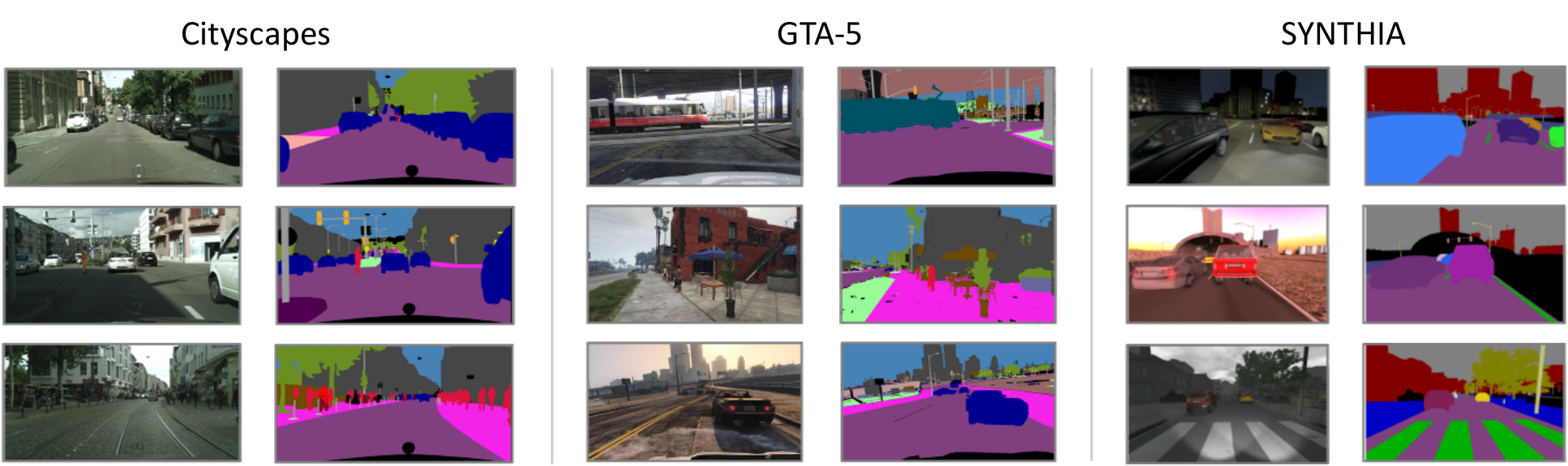}
\caption{
{\bf Left:} Samples from Cityscapes~\cite{CordtsCVPR16CityscapesDataset} recorded in the real world. They allow to evaluate the model performance on images that resemble the ones an agent will cope with at deployment; the difficulty of collecting real, large-scale datasets is the huge cost required to obtain fine annotations.
{\bf Middle:} Synthetic data from GTA-5~\cite{RichterECCV16PlayingForData}, 
obtained with high quality game engines, which makes easy the  pixel-wise annotation for SiS
and scene understanding. However, if the domain shift between real and synthetic data is not addressed, models trained on GTA-5 perform poorly on Cityscapes. 
{\bf Right:} An autonomous car must cope with large variations, such as day vs. night, weather condition changes, or structural differences, which might affect the image appearance even when the image is taken from the same viewpoint. Simulation engines allow generating large number of samples from urban environments in different conditions, as for example in the SYNTHIA~\cite{RosCVPR16SYNTHIADataset} dataset.}
\label{fig:segmDAex}
\end{center}
\end{figure*}

\begin{figure*}[ttt]
\begin{center}
\includegraphics[width=0.9\textwidth]{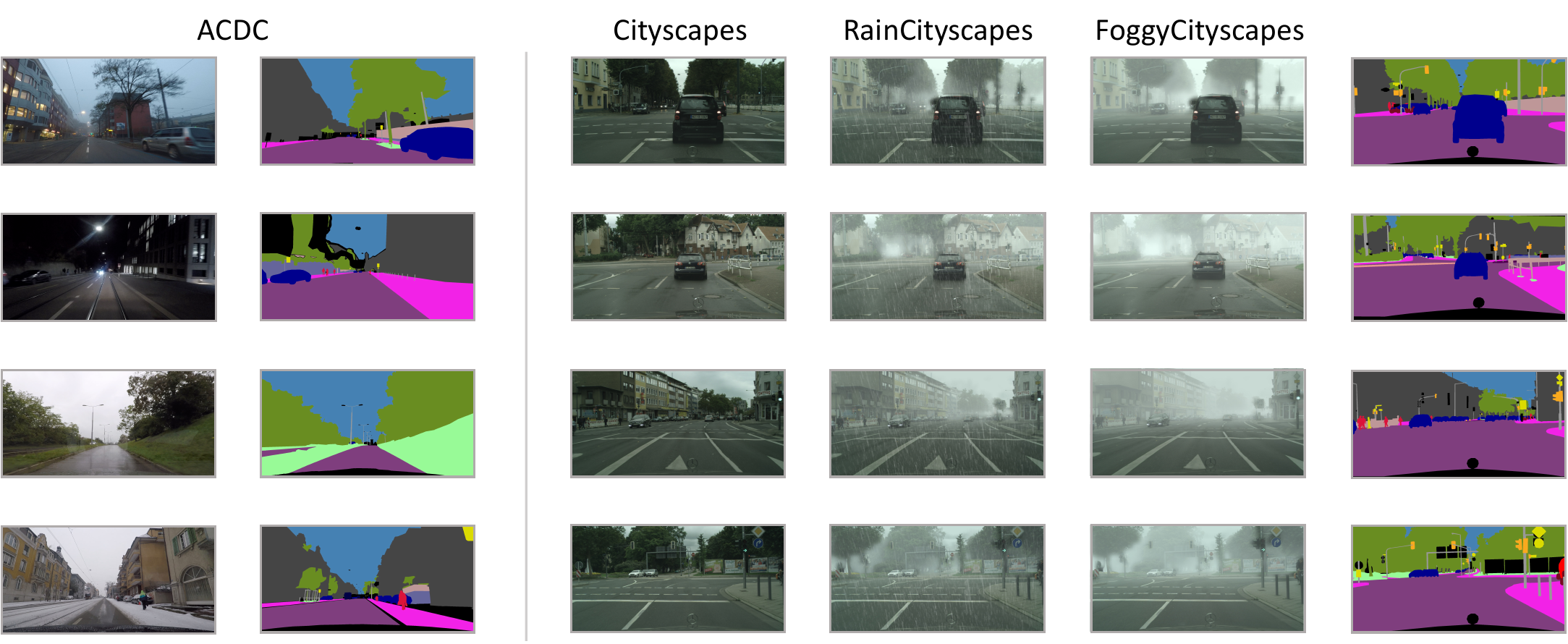}
\caption{
{\bf Left:} Example images from ACDC dataset~\cite{Sakaridis21ACDCDataset}, which permits to assess the model performance on real-world weather condition changes (fog, night, snow, rain).
{\bf Right:} Example images from RainCityscape~\cite{HuCVPR2019DepthAttentionalFeaturesSingleImageRainRemoval} and
FoggyCityscape~\cite{SakaridisECCV2018ModelAdaptSynthRealDataSemDenseFoggySceneUnderstanding}, which provide Cityscapes~\cite{CordtsCVPR16CityscapesDataset} images with simulated rain and fog, respectively.}
\label{fig:ACDC}
\end{center}
\end{figure*}

\section{Semantic image segmentation benchmarks}
\label{sec:benchmarks}

Understanding traffic scene images taken from vehicle mounted cameras is important for such advanced tasks as autonomous driving and driver assistance. It is a challenging problem due to large variations under different weather or illumination conditions
~\cite{DiTITS18CrossDomainTrafficSceneUnderstandingTL} or when a model needs to cope with different environments such as city, countryside and highway. One of the most used datasets for SiS of urban scenes is Cityscapes~\cite{CordtsCVPR16CityscapesDataset} (see examples in \fig{segmDAex} (left)); it comprises sequences of \textit{real and annotated} samples from a variety of European cities (mostly in Germany, Switzerland and Austria).

Even though relying on real samples allows assessing model performance in conditions that are more similar to deployment ones, manually annotating an image at pixel level for SiS 
is a very tedious and costly operation.
Recent progresses in computer graphics and modern graphics platforms such as game engines raise the prospect of easily obtaining labeled, synthetic datasets. Some examples in this directions are SYNTHIA~\cite{RosCVPR16SYNTHIADataset}, and GTA-5~\cite{RichterECCV16PlayingForData} 
(see examples in~\fig{segmDAex} (the middle and right side)).

However, models learned on such datasets might not be optimal due to the domain shift between synthetic and real data. To tackle this problem, a large set of DASiS methods have been proposed, most of which we surveyed in Sections~\ref{sec:dass} and~\ref{sec:beyond_dass}.
These methods start with a model pre-trained on the  
simulated data (typically GTA-5 or SYNTHIA) which is adapted
to real data, for which we assume not being able to access ground truth annotations (generally relying on Cityscapes~\cite{CordtsCVPR16CityscapesDataset}). 
This scenario mimics the realistic conditions such that a large database of simulated, labelled samples is available, and we intend to adapt the model to the real world with no access to ground truth annotations.

\begin{table*}
\begin{center}
{
\small
\begin{tabular}{@{}lcccccc@{}}
\multicolumn{7}{c}{\textbf{SiS datasets}} \\
\toprule


\textbf{Dataset's name}
& \textbf{\# Classes}
& \begin{tabular}{@{}c@{}} \textbf{\# Annotated} \\ \textbf{samples} \end{tabular}
& \begin{tabular}{@{}c@{}} \textbf{Real or} \\ \textbf{simulated} \end{tabular}
& \begin{tabular}{@{}c@{}} \textbf{Video} \\ \textbf{sequences} \end{tabular}
& \begin{tabular}{@{}c@{}} \textbf{Environment/} \\ \textbf{geography} \end{tabular}
& \begin{tabular}{@{}c@{}} \textbf{Controlled visual} \\ \textbf{conditions} \end{tabular} \\

\midrule
\midrule

Cityscapes~\cite{CordtsCVPR16CityscapesDataset} & $30$ & $5,000$* & Real & Yes & Germany; Zurich & $-$ \\
\midrule
BDD100K~\cite{YuCVPR2020BDD100KDataset} & $19$ & $10,000$ & Real & No & United States & $-$ \\
\midrule
KITTI~\cite{GeigerCVPR2021AreWeReadyADKITTIVisionBenchmarkSuite} & $28$ & $400$ & Real & Yes & Germany & $-$ \\
\midrule
CamVid~\cite{BrostowPRL2009SemObjClassesVideoHighDefGTDatabase} & $32$ & $701$ & Real & Yes & Cambridge (UK) & $-$ \\
\midrule
Mapillary~\cite{NeuholdICCV2017MapillaryDataset} & $66$ & $25,000$ & Real & No & Worldwide & $-$ \\
\midrule
IDD~\cite{VarmaWACV19IDDDatasetExploringADUnconstrainedEnvironments} & $34$ & $10,004$ & Real & Yes & India & $-$ \\
\midrule
RainCityscapes~\cite{HuCVPR2019DepthAttentionalFeaturesSingleImageRainRemoval} & $32$ & $10,620$ & Real & Yes & Germany & Artificial rain \\
\midrule
FoggyCityscapes~\cite{SakaridisECCV2018ModelAdaptSynthRealDataSemDenseFoggySceneUnderstanding} & $32$ & $15,000
$ & Real & Yes & Germany & Artificial fog \\
\midrule
ACDC~\cite{Sakaridis21ACDCDataset} & 19 & $4,006$ & Real & Yes & Switzerland & \begin{tabular}{@{}c@{}}Fog; Night; \\ Rain; Snow \end{tabular}  \\
\midrule
FoggyZurich~\cite{SakaridisECCV2018ModelAdaptSynthRealDataSemDenseFoggySceneUnderstanding} &
$19$ & $40$** 
& Real & Yes & Zurich & Fog \\
\midrule
GTA-5~\cite{RichterECCV16PlayingForData} & $19$ & $24,966$ & Sim & Yes & $-$ & $-$ \\
\midrule
SYNTHIA~\cite{RosCVPR16SYNTHIADataset} & $13$ & $200,000$ & Sim &  Yes & \begin{tabular}{@{}c@{}} Highway; NYC; \\ Europe \end{tabular} & \begin{tabular}{@{}c@{}} Season; Daytime \\ Fog; (Soft) Rain \end{tabular} \\
\midrule
SYNTHIA-RAND~\cite{RosCVPR16SYNTHIADataset} & $11$ & $13,407$ & Sim & No & $-$ & $-$ \\
\midrule
KITTI-v2~\cite{CabonX20VirtualKITTI2} & $15$ & $21,260$ & Sim & Yes & Germany & \begin{tabular}{@{}c@{}} Daytime; Overcast; \\ Fog; Rain \end{tabular} \\
\midrule
Synscapes~\cite{Wrenninge18SynscapesDataset} & $19$ & $25,000$ & Sim & No & - & \begin{tabular}{@{}c@{}} Daytime; Overcast; \\ Scene param. \end{tabular} \\

\bottomrule

\end{tabular}
} 
\end{center}
\caption{Datasets for research on urban scene SiS. From leftmost to rightmost columns, we indicate the dataset's name, the number of categories covered by annotations, whether the dataset contains real or simulated (rendered) images, whether samples are recorded as video sequences (\ie, mimicking an agent driving in an urban environment), the locationz from which samples were recorded or the ones simulated by the engine, and whether different visual conditions can be set by the user.
*Cityscapes~\cite{CordtsCVPR16CityscapesDataset} also contains $20,000$ additional samples with coarse annotations.
**FoggyZurich~\cite{SakaridisECCV2018ModelAdaptSynthRealDataSemDenseFoggySceneUnderstanding} also contains $3,808$ unlabelled samples. 
}
\label{tab:semsegm_datasets}
\end{table*}

While GTA-5, SYNTHIA and Cityscapes are the most widely used datasets to train and evaluate DASiS approaches, there exists a large body of (public) datasets for SiS of urban scenes available in the literature. In Table~\ref{tab:semsegm_datasets} we provide a summary of SiS datasets we are aware of, reporting their most important characteristics:
the number of classes, the number of annotated samples, whether images are real or rendered, whether the dataset contains video sequences (and not temporally uncorrelated images), the geographical location (for what concerns simulated datasets, we report the simulated area indicated, if available), and whether the dataset allows setting arbitrary conditions (seasonal, weather, daylight, \etc).
In addition, in Table~\ref{tab:semsegm_classes} we present a summary of the classes available in the different datasets, to ease the comprehension of the compatibility between different models --- also in light of new domain adaptation problems (shortly discussed in \ref{ssec:universal}).


\begin{table*}
\begin{center}
{
\small 
\begin{NiceTabular}{@{}|l|ccccccccccc|@{}}[code-before = 
\rowcolor{gray!10}{3}
\rowcolor{gray!10}{5}
\rowcolor{gray!10}{7}
\rowcolor{gray!10}{9}
\rowcolor{gray!10}{11}
\rowcolor{gray!10}{13}
\rowcolor{gray!10}{15}
\rowcolor{gray!10}{17}
\rowcolor{gray!10}{19}
\rowcolor{gray!10}{21}
\rowcolor{gray!10}{23}
\rowcolor{gray!10}{25}
\rowcolor{gray!10}{27}
\rowcolor{gray!10}{29}
]
\toprule


  \rotatebox{0}{ \textbf{Classes} }
& \rotatebox{75}{ Cityscapes~\cite{CordtsCVPR16CityscapesDataset} }
& \rotatebox{75}{ BDD100K~\cite{YuCVPR2020BDD100KDataset} }
& \rotatebox{75}{ CamVid~\cite{BrostowPRL2009SemObjClassesVideoHighDefGTDatabase} }
& \rotatebox{75}{ IDD~\cite{VarmaWACV19IDDDatasetExploringADUnconstrainedEnvironments} }
& \rotatebox{75}{ ACDC~\cite{Sakaridis21ACDCDataset} }
& \rotatebox{75}{ GTA-5~\cite{RichterECCV16PlayingForData} }
& \rotatebox{75}{ SYNTHIA-RAND~\cite{RosCVPR16SYNTHIADataset} }
& \rotatebox{75}{ SYNTHIA~\cite{RosCVPR16SYNTHIADataset} }
& \rotatebox{75}{ KITTI-v2~\cite{CabonX20VirtualKITTI2} }
& \rotatebox{75}{ FoggyZurich~\cite{SakaridisECCV2018ModelAdaptSynthRealDataSemDenseFoggySceneUnderstanding} }
& \rotatebox{75}{ Synscapes~\cite{Wrenninge18SynscapesDataset} }
\\

\midrule
\midrule

Animal               &            &            & \checkmark & \checkmark &            &            &            &            &            &            &            \\
%
Bicycle              & \checkmark & \checkmark & \checkmark & \checkmark & \checkmark & \checkmark & \checkmark & \checkmark &            & \checkmark & \checkmark \\
Bridge               & \checkmark &            & \checkmark & \checkmark &            &            &            &            &            &            &            \\
Building             & \checkmark & \checkmark & \checkmark & \checkmark & \checkmark & \checkmark & \checkmark & \checkmark & \checkmark & \checkmark & \checkmark \\
Bus                  & \checkmark & \checkmark & \checkmark & \checkmark & \checkmark & \checkmark & \checkmark &            &            & \checkmark & \checkmark \\
Car                  & \checkmark & \checkmark & \checkmark & \checkmark & \checkmark & \checkmark & \checkmark & \checkmark & \checkmark & \checkmark & \checkmark \\
Caravan              & \checkmark &            &            & \checkmark &            &            &            &            & \checkmark &            &            \\
%
Fence                & \checkmark & \checkmark & \checkmark & \checkmark & \checkmark & \checkmark & \checkmark & \checkmark &            & \checkmark & \checkmark \\
Guard rail           & \checkmark &            &            & \checkmark &            &            &            &            & \checkmark &            &            \\
Lane marking         &            &            & \checkmark &            &            &            & \checkmark & \checkmark &            &            &            \\
Motorcycle           & \checkmark & \checkmark & \checkmark & \checkmark & \checkmark & \checkmark & \checkmark &            &            & \checkmark & \checkmark \\
%
Parking              & \checkmark &            & \checkmark & \checkmark &            &            & \checkmark &            &            &            &            \\
Person               & \checkmark & \checkmark & \checkmark & \checkmark & \checkmark & \checkmark & \checkmark & \checkmark &            & \checkmark & \checkmark \\
Pole                 & \checkmark & \checkmark & \checkmark & \checkmark & \checkmark & \checkmark & \checkmark & \checkmark & \checkmark & \checkmark & \checkmark \\
Rail track           & \checkmark &            &            & \checkmark &            &            &            &            &            &            &            \\
Rider                & \checkmark & \checkmark &            & \checkmark & \checkmark & \checkmark & \checkmark &            &            & \checkmark & \checkmark \\
Road                 & \checkmark & \checkmark & \checkmark & \checkmark & \checkmark & \checkmark & \checkmark & \checkmark & \checkmark & \checkmark & \checkmark \\
%
Sky                  & \checkmark & \checkmark & \checkmark & \checkmark & \checkmark & \checkmark & \checkmark & \checkmark & \checkmark & \checkmark & \checkmark \\
Sidewalk             & \checkmark & \checkmark & \checkmark & \checkmark & \checkmark & \checkmark & \checkmark & \checkmark &            & \checkmark & \checkmark \\
Terrain              & \checkmark & \checkmark &            &            & \checkmark & \checkmark & \checkmark &            & \checkmark & \checkmark & \checkmark \\
Trailer              & \checkmark &            &            & \checkmark &            &            &            &            &            &            &            \\
Train                & \checkmark & \checkmark & \checkmark &            & \checkmark & \checkmark & \checkmark &            &            & \checkmark & \checkmark \\
Traffic light        & \checkmark & \checkmark & \checkmark & \checkmark & \checkmark & \checkmark & \checkmark &            & \checkmark & \checkmark & \checkmark \\
Traffic sign         & \checkmark & \checkmark & \checkmark & \checkmark & \checkmark & \checkmark & \checkmark & \checkmark & \checkmark & \checkmark & \checkmark \\
Tree                 &            &            & \checkmark &            &            &            &            &            & \checkmark &            &            \\
Truck                & \checkmark & \checkmark & \checkmark & \checkmark & \checkmark & \checkmark & \checkmark &            & \checkmark & \checkmark & \checkmark \\
Tunnel               & \checkmark &            & \checkmark & \checkmark &            &            &            &            &            &            &            \\
Vegetation           & \checkmark & \checkmark & \checkmark & \checkmark & \checkmark & \checkmark & \checkmark & \checkmark & \checkmark & \checkmark & \checkmark \\
Wall                 & \checkmark & \checkmark & \checkmark & \checkmark & \checkmark & \checkmark & \checkmark &            &            & \checkmark & \checkmark \\

\bottomrule

\end{NiceTabular}
} 
\end{center}
\caption{Categories of which annotation is provided in different SiS datasets. We report classes available in at least
two distinct datasets: some datasets, such as CamVid~\cite{BrostowPRL2009SemObjClassesVideoHighDefGTDatabase}, contain a variety
of other categories; we refer to the references for their descriptions.} 
\label{tab:semsegm_classes}
\end{table*}


\myparagraph{DASiS benchmarks}
The most common settings used in the DASiS research is summarized in Table~\ref{tab:semsegm_benchmarks}. They were introduced in the pioneering DASiS study by Hoffman~\etal~\cite{HoffmanX16FCNsInTheWildPixelLevelAdversarialDA}. As the first row in the table indicates, the most widely used benchmark is GTA-5~\cite{RichterECCV16PlayingForData} $\rightarrow$ Cityscapes~\cite{CordtsCVPR16CityscapesDataset} task.
It represents a sim-to-real adaptation problem, since GTA-5~\cite{RichterECCV16PlayingForData} was conceived to be consistent with Cityscapes~\cite{CordtsCVPR16CityscapesDataset} annotations. Following the notation from Section~\ref{sec:dass}, the source dataset $\calD_\calS$ is defined by GTA-5~\cite{RichterECCV16PlayingForData} annotated samples, and the target dataset $\calD_\calT$ is defined by Cityscapes~\cite{CordtsCVPR16CityscapesDataset} (non-annotated) samples. 

 \begin{table}[ttt]
\begin{center}
{
\footnotesize 
\begin{tabular}{@{}lcc@{}}
\multicolumn{3}{c}{\textbf{Main benchmarks for DASiS}} \\
\toprule

  \begin{tabular}{@{}c@{}} \textbf{Source} \\ \textbf{domain} \end{tabular}
& \begin{tabular}{@{}c@{}} \textbf{Target} \\ \textbf{domain} \end{tabular} 
& \begin{tabular}{@{}c@{}} \textbf{Adaptation} \\ \textbf{type} \end{tabular}  \\

\midrule
\midrule

  GTA-5~\cite{RichterECCV16PlayingForData}
& Cityscapes~\cite{CordtsCVPR16CityscapesDataset}
& Sim-to-real \\
\midrule

  SYNTHIA-RAND~\cite{RosCVPR16SYNTHIADataset}
& Cityscapes~\cite{CordtsCVPR16CityscapesDataset}
& Sim-to-real \\
\midrule

  Cityscapes~\cite{CordtsCVPR16CityscapesDataset} (Train)
& Cityscapes~\cite{CordtsCVPR16CityscapesDataset} (Val)
& Cross-city (real) \\
\midrule

  SYNTHIA~\cite{RosCVPR16SYNTHIADataset} (Fall)
& SYNTHIA~\cite{RosCVPR16SYNTHIADataset} (Winter)
& Cross-weather (sim) \\


\bottomrule

\end{tabular}
} 
\end{center}
\caption{The most widely used benchmarks within the DASiS community. The first column 
indicates the source dataset (labeled images available); the second column 
indicates the target dataset (unlabeled images available); the third column indicates the
type of adaptation problem.} 
\label{tab:semsegm_benchmarks}
\end{table}


Naturally, datasets generated with the help of simulation engines are significantly larger, as they are able to generate synthetic data under a broad set of conditions (the only exception is GTA-5~\cite{RichterECCV16PlayingForData}, that is considerably large but does not allow the user to set different visual conditions). Still, in order to evaluate how the models will perform in the real environment on various real conditions, these synthetic  datasets  might be not sufficient.
Therefore, an important contribution to the semantic segmentation landscape is the real-image ACDC dataset~\cite{Sakaridis21ACDCDataset}, that is both reasonably large (slightly smaller than Cityscapes~\cite{CordtsCVPR16CityscapesDataset}) and flexible in terms of visual conditions: researchers can indeed choose between foggy, dark, rainy and snowy scenarios --- and, importantly, samples are recorded from the same streets in such different conditions, allowing to properly assess the impact of adverse weather/daylight (see examples in \fig{ACDC} (left)).  RainCityscape~\cite{HuCVPR2019DepthAttentionalFeaturesSingleImageRainRemoval} and FoggyCityscape~\cite{SakaridisECCV2018ModelAdaptSynthRealDataSemDenseFoggySceneUnderstanding} are also extremely valuable in this direction, but in this case the weather conditions are simulated (on top of the real Cityscapes images). We think that these datasets are better suited than the currently used Cityscapes dataset and we expect that in the future DASiS methods will be also evaluated on these or similar datasets (see examples in \fig{ACDC} (right)). 
 
\subsection{Evaluating SiS and DASiS}
\label{sec:evaluate}

In this section, we first recall the main measures proposed in the literature to evaluate the quality of semantic image segmentation. 
Then, we provide a short summary of evaluation protocols that are currently used in DA/DASiS and discuss some of their limitations.

\myparagraph{SiS evaluation measures}
To evaluate SiS, the overall pixel accuracy, which measures the proportion of correctly labelled pixels and the per-class accuracy measures the proportion of correctly labelled pixels for each class and then averages over the classes have been proposed in \cite{ShottonECCV06TextonBoostJointAppearanceShapeContextSemSegm}. The Jaccard Index (JI) or more popularly known intersection over the union (IoU) takes into account both the false 
positives
and the missed values for each class and can be derived from the  confusion matrix. It measures 
the intersection over the union of the labelled segments for each class and reports the average; it became the standard to evaluate SiS models, after having been introduced in the Pascal VoC challenge \cite{EveringhamIJCV10ThePascalVisualObjectClassesVOCChallenge} in 2008. 
Long~\etal~\cite{LongCVPR15FullyConvolutionalNetworksSegmentation} propose in addition a frequency weighted IoU measure where the IoU for each class is weighted by the frequency of GT pixels corresponding to that class. 
 
We schematize such main metrics below, following notation used by Long~\etal~\cite{LongCVPR15FullyConvolutionalNetworksSegmentation}. 
Let $n_{ij}$ be the number of pixels from the $i_{th}$ class that are classified as belonging to the $j_{th}$ class; let $n_{cl}$ be the number of different classes; let $t_i = \sum_j n_{ij}$ be the total number of pixels of the $i_{th}$ class. The metrics introduces above are defined as follows

\begin{itemize}
  \item \textbf{Mean IoU}: 
    $\frac{1}{n_{cl}}\sum_i\frac{ n_{ii}}{(t_i + \sum_j n_{ji} - n_{ii})}$
  \item \textbf{Frequency weighted IoU}:
    $\frac{1}{\sum_k t_k}\sum_i\frac{t_i\cdot n_{ii}}{(t_i + \sum_j n_{ji} - n_{ii})}$
  \item \textbf{Pixel accuracy}: 
    $\frac{\sum_i n_{ii}}{\sum_i t_i}$
  \item \textbf{Mean accuracy}: 
    $\frac{1}{n_{cl}}\sum_i\frac{ n_{ii}}{t_i}.$
\end{itemize}

The above measures are derived in  general from the confusion matrix computed over the whole dataset, which has the main advantage that there is no need to handle the absent classes in each image. While these metrics are the most used to evaluate and compare DASiS models, we would like to mention below a few other metrics that have been introduced  in the literature to evaluate SiS models in general, and hence they could be interesting  also for evaluating DASiS.

Instead of relying on the confusion matrix computed over the whole dataset, 
Csurka~\etal~\cite{CsurkaBMVC13WhatGoodEvalMeasureSemSegm} propose to evaluate the pixel accuracy, the mean  accuracy and  the IoU for each image individually, where  the IoU is  computed by  averaging over only  the classes present in the GT segmentation of the image. The main rationale behind this is 
that measures computed over the whole dataset do not enable to distinguish an algorithm that delivers a medium score on all images from an algorithm that performs very well on some images and very poorly on others (they could yield  a very similar average). To better assess such differences, Furthermore, Csurka~\etal~\cite{CsurkaBMVC13WhatGoodEvalMeasureSemSegm} propose to measure the percentage of images with a performance higher than a given threshold  and given a pair of approaches, the percentage of images for which one of the method outperforms the other one 
to analyze the statistical difference of two segmentation algorithms with t-test. Finally, it has also been noticed in~\cite{CsurkaBMVC13WhatGoodEvalMeasureSemSegm}, that  per-image scores reduce the bias \wrt large objects, as missing or incorrectly segmented small objects have a lower impact on the global confusion matrix.

Another important aspect of semantic segmentation is an accurate semantic border detection. To evaluate the accuracy of boundary segmentation, Kohli~\etal~\cite{KohliIJCV09RobustHigherOrderPotentialsEnforcingLabelConsistency} propose Trimap that defines a narrow band around each contour and computes pixel accuracies in the given band;  Csurka~\etal~\cite{CsurkaBMVC13WhatGoodEvalMeasureSemSegm} extend the Berkeley contour matching (BCM) score~\cite{MartinPAMI04LearningToDetectNaturalImageBoundaries} --- proposed to evaluate similarity between unsupervised segmentation and human annotations --- to SiS where a BCM is computed between the GT and predicted contours corresponding to each class.

\myparagraph{DA and DASiS evaluation protocols} 
There exist two main evaluation protocols in DA: 
\textit{transductive} and \textit{inductive}. Transductive DA aims to learn prediction models that directly assign labels to the target instances available during training. In other words, the model aims to perform well on the sample set $\calD_\calT$ used to learn the model. Instead, the inductive UDA measures the performance of the learned models on held-out target instances that are sampled from the same target distribution, $\hat{\calD_\calT} \sim P_\calT$.
While in classical DA most often the transductive protocol is considered, in the case of DASiS, the \textit{inductive} setting is the preferred one.

Selecting the best models, hyperparameter settings is rather challenging in practice. As described in \cite{SaitoICCV21TuneItTheRightWayUnsupervisedValidationDASoftNeighborhoodDensity}, many methods do hyper-parameter optimization using the risk on the target domain, which contradicts the core assumption of UDA; in many papers, a clear description about how the final model has been selected for evaluation is often missing, making the comparisons between different methods rather questionable. 
Even if in the inductive evaluation protocol a different set is used to select the model, an obvious question arises: \textit{If the model has access to target labels for evaluation, why not using those labelled target samples to improve the model in a semi-supervised DA fashion?}

Fairer strategies such as transfer cross-validation \cite{ZhongECML10CrossValidationTL}, reverse cross-validation \cite{GaninJMLR16DomainAdversarialNN}, importance-weighted cross-validation~\cite{LongNIPS18ConditionalAdversarialDomainAdaptation}, deep embedded validation~\cite{YouICML19TowardsAccurateModelSelectionDA} rely on source labels, evaluating the risk in the source domain and/or exploiting the data distributions. However, these strategies remain sub-optimal due to the fact that they still rely on 
the source risk which is not necessarily a good
estimator of the target risk in the presence of a large domain gap \cite{SaitoICCV21TuneItTheRightWayUnsupervisedValidationDASoftNeighborhoodDensity}.

Instead, Saito~\etal~\cite{SaitoICCV21TuneItTheRightWayUnsupervisedValidationDASoftNeighborhoodDensity}  revisit the unsupervised validation criterion based on the classifier entropy and show that when the classification model produces confident and low-entropy outputs on target samples  the target features are discriminative and the predictions likely reliable. However, they claim that such criterion is unable to detect when a DA method falsely align target samples with the source and incorrectly changes the neighborhood structure. To overcome this limitation, they propose a model selection method based on  soft neighborhood density measure to evaluate the discriminability of target features.




\section{Conclusions}\label{sec:conclusions}


In this paper, we provide a comprehensive and up-to-date review of the domain adaptation of semantic image segmentation (DASiS) literature. We describe main trends and organize DASiS
methods according to their most important characteristics --- such as the backbone segmentation network, the type and levels of alignment, 
parameter
sharing etc. We complement the survey with a brief recall of the main deep methods proposed to tackle the task of semantic image segmentation (SiS), and we extend DASiS with some newer methods,
where the assumptions that both source and target domains consist of samples drawn from single data distributions and that they are both available are relaxed, discussing  and surveying proposed methods for tasks such as multi-source or multi-target DA, domain generalization and source-free adaptation. Finally, we provide an extensive comparison of the existing SiS datasets and discuss different evaluation measures and protocols, allowing to compare different approaches. 


As the survey shows, since the very first work on the topic~\cite{HoffmanX16FCNsInTheWildPixelLevelAdversarialDA}, DASiS has been a very active research field, with an increasing number of approaches being developed by the community and actively integrated in advanced industrial applications and solutions for autonomous driving, robot navigation, medical imaging, remote sensing, \etc
Therefore, we believe that the community can 
benefit from our survey, in particular, PhD students and young researchers who are just beginning their work in the domain, but also developers from the industry willing to integrate DASiS in their systems can find answers to their questions. 



{\small
\bibliographystyle{ieee_fullname}
\bibliography{egbib}
}

\end{document}